\documentclass[10pt,journal,compsoc]{IEEEtran}

\usepackage{epsfig}
\usepackage{graphicx}
\usepackage{amsmath, amsfonts}
\usepackage{amssymb}

\usepackage{float}
\usepackage{booktabs}
\usepackage{caption}
\usepackage{subcaption}
\usepackage{multirow}
\usepackage{array}
\usepackage{xcolor}
\usepackage{cite}
\usepackage[utf8]{inputenc}
\usepackage[vlined,ruled]{algorithm2e}
\usepackage{cuted}
\usepackage[pagebackref=true,breaklinks=true,colorlinks,bookmarks=false]{hyperref}
\usepackage{url}
\usepackage{ragged2e}
\usepackage{siunitx}
\usepackage{tabularray} 
\usepackage{pifont} 
\usepackage{arydshln}
\usepackage[percent]{overpic} 
\usepackage{xcolor}
\newcommand{\overlaybold}[1]{%
  \mbox{%
    \makebox[0pt][l]{#1}
    \makebox[0pt][l]{\raisebox{0.15pt}{#1}}
    \makebox[0pt][l]{\raisebox{-0.15pt}{#1}}
    #1
  }%
}

\newcommand{\cmark}{\ding{51}} 
\newcommand{\xmark}{\ding{55}} 

\makeatletter
\DeclareRobustCommand\onedot{\futurelet\@let@token\@onedot}
\def\@onedot{\ifx\@let@token.\else.\null\fi\xspace}

\def\etal{\emph{et al}\onedot}
\makeatother

\begin{document}
%
\title{Text2Loc++: Generalizing 3D Point Cloud Localization from Natural Language}


\author{Yan~Xia\textsuperscript{\dag}*,
        Letian~Shi*,
        Yilin~Di,
        João~F.~Henriques,
        and~Daniel~Cremers 
\thanks{\textsuperscript{\dag} Yan Xia is the corresponding author of this work. * Yan Xia and Letian Shi contributed equally to this work.\\
Yan Xia is with the School of Artificial Intelligence and Data Science, University of Science and Technology of China, 230026 Hefei, China (e-mail: yan.xia@ustc.edu.cn). \\
Letian Shi, Yilin Di, and Daniel Cremers are with Technical University of Munich, 80333 Munich, Germany (e-mail: letian.shi@tum.de, yilin.di@tum.de, cremers@tum.de). \\
Letian Shi is also with Institute for Computer Science, Artificial Intelligence and Technology (INSAIT), 1784, Bulgaria.\\
João~F.~Henriques is with the Visual Geometry Group, University of Oxford, Oxford OX1 2JD, United Kingdom (e-mail: joao@robots.ox.ac.uk). 
}%
}

%
%

\markboth{Journal of \LaTeX\ Class Files,~Vol.~14, No.~8, August~2015}%
{Shell \MakeLowercase{\textit{et al.}}: Bare Advanced Demo of IEEEtran.cls for IEEE Computer Society Journals}
%



\IEEEtitleabstractindextext{%
\begin{justify} 
\begin{abstract}
We tackle the problem of localizing 3D point cloud submaps using complex and diverse natural language descriptions, and present Text2Loc++, a novel neural network designed for effective cross-modal alignment between language and point clouds in a coarse-to-fine localization pipeline. To support benchmarking, we introduce a new city-scale dataset covering both color and non-color point clouds from diverse urban scenes, and organize location descriptions into three levels of linguistic complexity. In the global place recognition stage, Text2Loc++ combines a pretrained language model with a Hierarchical Transformer with Max Pooling (HTM) for sentence-level semantics, and employs an attention-based point cloud encoder for spatial understanding. We further propose Masked Instance Training (MIT) to filter out non-aligned objects and improve multimodal robustness. To enhance the embedding space, we introduce Modality-aware Contrastive Learning (MCL), incorporating cross-modal, submap-, text-, and instance-level losses. In the fine localization stage, we completely remove explicit text-instance matching and design a lightweight yet powerful framework based on Prototype-based Map Cloning (PMC) and a Cascaded Cross-Attention Transformer (CCAT). Extensive experiments on the KITTI360Pose dataset show that Text2Loc++ outperforms existing methods by up to 15\%. In addition, the proposed model exhibits robust generalization when evaluated on the new dataset, effectively handling complex linguistic expressions and a wide variety of urban environments.

\end{abstract}
\end{justify}

}

\maketitle

\IEEEdisplaynontitleabstractindextext

%
\IEEEpeerreviewmaketitle

\begin{figure*}[h]
\centering
\includegraphics[width=0.99\linewidth]{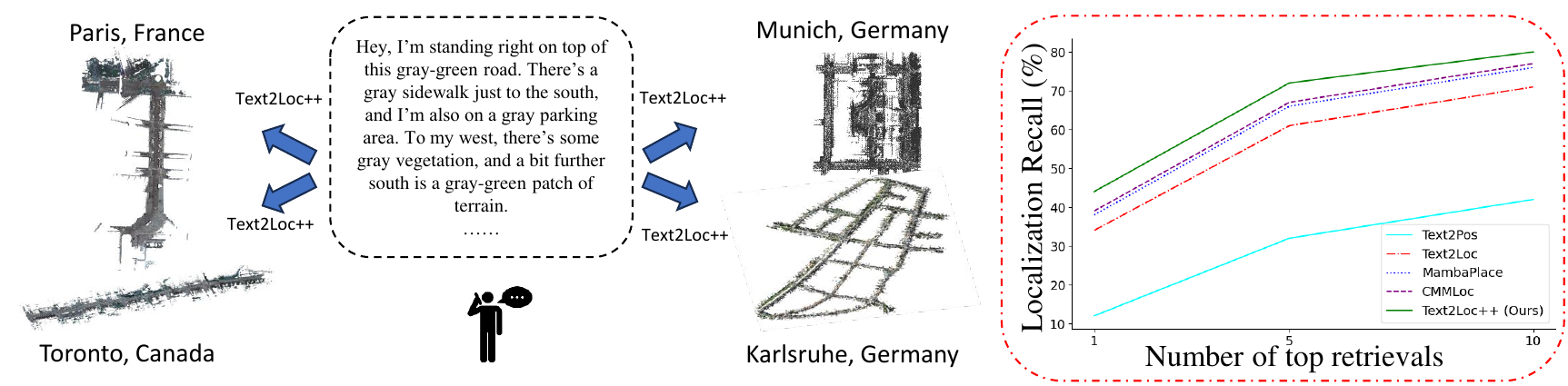}
\captionof{figure}{\textit{(Left)} We present Text2Loc++, a framework developed for cross-city and cross-country position localization based on textual descriptions of varying complexity. Given a 3D point cloud representing the surrounding environment and a textual query describing a location from different urban or regional contexts, Text2Loc++ identifies the most probable position corresponding to the description within the map. The model demonstrates strong generalization across diverse language inputs and heterogeneous point cloud data.
\textit{(Right)} Localization results on the KITTI360Pose test set show that the proposed Text2Loc++ consistently outperforms existing baselines across all top-k retrieval thresholds, achieving up to 15\% higher accuracy in localizing text queries within a \SI{5}{m} error range. The textual query here is expressed in the \emph{complex} linguistic form.}
\label{fig:teaser}
\end{figure*}

\section{Introduction}
\label{sec:intro}
3D localization~\cite{Min_2023_CVPR, Sarlin_2023_CVPR} from natural language descriptions within large-scale urban maps plays a crucial role in enabling autonomous agents to collaborate effectively with humans for trajectory planning~\cite{hu2023_uniad}. Such capability is essential in real-world applications including goods delivery and vehicle pickup~\cite{xia2023lightweight, xia2021vpc}.
In practice, tasks such as food delivery often encounter the so-called "last mile problem". Identifying the exact drop-off point within residential neighborhoods or large office complexes is particularly challenging, as GPS signals tend to degrade or fail in environments surrounded by tall buildings and dense vegetation~\cite{Xia_2023_ICCV, xia2023perception}. Couriers commonly rely on verbal guidance from recipients to reach the correct destination. More generally, this “last mile problem” manifests whenever users attempt to navigate to unfamiliar or occluded locations, thereby highlighting the necessity of developing localization systems that can operate directly from natural language inputs which, as shown in Fig.~\ref{fig:teaser}.

A promising direction is to align linguistic descriptions with pre-constructed 3D point-cloud maps obtained from calibrated depth sensors such as LiDAR. Compared with image-based localization, point-cloud-based methods focus on geometric scene structures and thus offer significant advantages: they exhibit strong robustness to changes in illumination, weather, and season, whereas the same scene captured in images may vary drastically in appearance under different environmental conditions~\cite{wang2025, zhang2026}.

However, text-to-3D localization paradigm still faces four primary challenges: (1) Effectively extracting discriminative features from both modalities to facilitate cross-modal tasks. (2) Bridging the structural gap for seamless alignment and fusion. (3) Addressing semantic misalignment and incomplete correspondences between descriptions and 3D submaps. (4) Overcoming the limited diversity in existing datasets regarding both textual styles and point cloud distributions to ensure robust generalization.

\begin{figure*}[t!]
  \centering
\includegraphics[width=0.99\textwidth]
{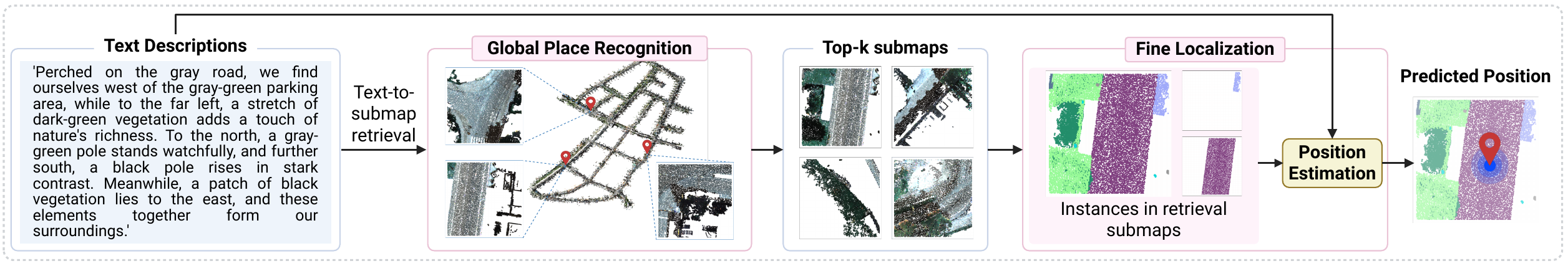}
\caption{The proposed Text2Loc++ architecture. It consists of two consecutive modules: Global place recognition and Fine localization. \textit{Global place recognition.} Given a text-based position description, we first identify a set of coarse candidate locations, "submaps," potentially containing the target position. This is achieved by retrieving the top-k nearest submaps from a previously constructed database of submaps using our novel text-to-submap retrieval model. \textit{Fine localization.} We then refine the center coordinates of the retrieved submaps via our designed matching-free position estimation module, which adjusts the target location to increase accuracy. The text description shown here is expressed in the \emph{complex} linguistic form.
}
\vspace{-1em}
  \label{fig:pipeline}
\end{figure*}

Fig.~\ref{fig:pipeline} shows our proposed text-to-3D localization framework, Text2Loc++, addressing these challenges via a robust two-stage pipeline comprising global recognition and fine localization. To address the challenge (1), we design dedicated encoders for text and 3D point clouds. On the linguistic side, we leverage the hierarchical nature of descriptions (sentences and tokens) by utilizing a frozen T5~\cite{2020t5} backbone and a hierarchical Transformer with max-pooling to capture intra- and inter-textual dependencies. On the point cloud side, we focus on the relational dynamics of 3D instances to enhance representation discriminability. Specifically, we advance the Text2Pose~\cite{kolmet2022text2pos} instance encoder by integrating a number encoder and an attention-driven fusion layer. These enhancements ensure more discriminative and comprehensive feature representations, thereby establishing a robust foundation for subsequent cross-modal alignment and fusion.

Specifically targeting the cross-modal alignment issue in challenge (2), we apply contrastive learning on both sentence-to-object level (instance loss) and paragraph-to-scene level (text-submap loss) to preserve balance between positive and negative pairs among different modalities. 
Meanwhile, the text–instance matching modules used in previous approaches can introduce severe noise due to inaccurate correspondence or offset predictions, thereby hindering precise position estimation. To mitigate this multi-modal fusion problem in challenge (2), we propose a novel matching-free fine localization network. Specifically, a prototype-based map cloning (PMC) module is designed to enrich the diversity of retrieved submaps, followed by a cascaded cross-attention transformer (CCAT) that integrates semantic information from point clouds into the text embedding.
Together, these components facilitate end-to-end training, allowing for the direct prediction of the target position without relying on any text–instance matcher. 

Challenge (3) represents a critical yet frequently overlooked bottleneck in current text-to-3D research. Similar to other multi-modal paradigms, text-to-point-cloud pairs are inherently noisy and characterized by a profound asymmetric correspondence~\cite{zhai2023sigmoid}. Specifically, a single 3D submap can be accurately depicted by multiple valid textual descriptions—each focusing on a distinct subset of spatial features—while, conversely, a single description may refer to multiple candidate point clouds with partially similar semantic layouts.
To handle this ambiguity, we utilize a Masked Instance Training strategy (MIT) to mask out non-aligned instances~\cite{yang2023attentivemaskclip} during training, ensuring that the model focuses only on text-relevant content. Inspired by~\cite{9578480} and~\cite{wang2023connecting}, we further enhance training by proposing submap-to-submap and text-to-text losses which encourages a more structured and semantically meaningful embedding space. We collectively refer to these constraints as Modality-aware Contrastive Learning (MCL). 
This diverse set of objectives significantly improves the robustness and generalization of the model without sacrificing accuracy.

Finally, we notice that the existing methods~\cite{kolmet2022text2pos, wang2023text, shang2025mambaplace, xu2025cmm} were primarily applied to simple and uniform text descriptions, where each sentence in the text typically conveys only a single piece of information, without the complexity or mixing found in everyday language.
In addition, although the training and testing sets of 3D point clouds come from different areas, they are derived from the same city, sharing similar annotation strategies and stylistic characteristics. Thus, it is unclear whether existing methods can maintain consistent effectiveness when dealing with more complex and diverse text data shown in Challenge (4). 

To this end, we construct a new city-scale text–point cloud localization dataset that includes both real and synthetic point clouds from diverse cities and countries: SemanticKITTI, Paris\_CARLA~\cite{deschaud2021pariscarla3d}, Toronto3D~\cite{tan2020toronto3d}, TUM City Campus~\cite{zhu2020tum}, and Paris\_Lille~\cite{roynard2017parislille3d}. All point clouds are annotated following the protocol of \cite{kolmet2022text2pos}. Besides, we generate varying levels of linguistic complexity while preserving semantic consistency to describe the locations, creating a more challenging and realistic benchmark for text–point cloud localization task.

To better handle complex and diverse textual descriptions, we introduce a new training strategy to enhance the model's language understanding and cross-modal alignment capabilities. Specifically, we adopt the LoRA approach~\cite{Hu2021LoRALA} to fine-tune the T5 encoder, enabling the model to extract semantic information from natural language inputs more effectively.
However, as text complexity increases, direct alignment of rich language descriptions with point clouds becomes increasingly challenging. Due to the incomplete alignment between the visible 3D content and the textual input, establishing accurate cross-modal correspondences is difficult, especially under a powerful yet unconstrained T5 encoder. To address this, we incorporate text distillation~\cite{sanh2020distilbert, xu2024}, which filters out irrelevant content and preserves only the most semantically salient information. This distilled representation improves the accuracy and robustness of multimodal matching.

To summarize, the main contributions of this work are:
\begin{itemize}

\item We propose a novel attention-based method that is hierarchical and represents contextual details within and across sentence descriptions of places. We also propose a novel attention-based point cloud module.

\item To address the ambiguity caused by non-aligned instances, we propose a Masked Instance Training (MIT) strategy that selectively filters out irrelevant objects during training in two stages.

\item To better organize the multimodal representation space, we introduce Modality-aware Contrastive Learning (MCL), incorporating cross-modal alignment, submap, text, and instance losses.

\item We are the first to completely remove the usage of the text-instance matcher in the final localization stage. We propose a lightweight and faster localization model that achieves state-of-the-art performance through our designed prototype-based map cloning (PMC) module in training and the cascaded cross-attention transformer (CCAT).

\item 
We introduce a new benchmark for text-to-point cloud localization 
with multi-level linguistic complexity and both color and non-color point clouds from diverse real-world cities. Through this benchmark, we are the first to empirically demonstrate the zero-shot generalization of text-to-3D localization models.

\end{itemize}
This journal article builds upon and extends our earlier conference paper presented at CVPR 2024~\cite{xia2024text2loc}. In this updated version, we introduce the following extension: 
I) We construct a new city-scale text–point cloud localization dataset using real and synthetic point clouds from diverse urban environments. Each location is described with semantically consistent texts of varying linguistic complexity, enabling more realistic and challenging benchmarking.
II) We propose a multiple
contrastive learning with a masked instance training strategy to address the challenge of partial matching, effectively reducing incorrect associations between modalities and boosting the robustness among different benchmarks. III) We employ LoRA and text distillation to address the challenges posed by increasing text complexity and diversity. 
IV) We conduct extensive experiments on the new benchmark and show that the proposed Text2Loc++ greatly improves over the state-of-the-art methods. Our code is at
\href{https://github.com/anslt/Text2LocPP}{https://github.com/anslt/Text2LocPP}.

\section{Related work}
In prior research, only a limited number of networks have been proposed for natural language–based 3D localization in large-scale outdoor environments. Closely related research areas include 2D visual localization, 3D point cloud–based localization, and language-grounded 3D understanding.

\textbf{2D Visual Localization.} Early systems utilized hand-crafted features like SIFT~\cite{lowe2004distinctive} and ORB~\cite{rublee2011orb} for viewpoint-invariant matching. Modern learning-based methods~\cite{sarlin2019coarse, sattler2016efficient} typically adopt a coarse-to-fine pipeline. In the coarse stage, global place recognition is performed via nearest-neighbor search in a high-dimensional embedding space to identify potential candidates. The fine stage then establishes dense, pixel-wise correspondences between the query and retrieved reference images to refine the pose estimation. However, image-based approaches remain sensitive to drastic appearance changes (e.g., lighting, weather). Unlike these unimodal visual tasks, our work tackles cross-modal 3D localization via textual queries.

\textbf{3D Point Cloud-based Localization.} Deep 3D localization often follows the same two-stage paradigm: place recognition and pose estimation. PointNetVlad~\cite{angelina2018pointnetvlad} pioneered end-to-end 3D recognition, followed by SOE-Net~\cite{xia2021soe}, which introduced orientation-aware local descriptors. To capture long-range dependencies, Transformer-based architectures~\cite{zhou2021ndt, deng2018ppfnet, fan2022svt, zhang2022rank, ma2022overlaptransformer, barros2022attdlnet, ma2023cvtnet, zheng2024regiformer} have been extensively explored. While voxel-based methods like MinkLoc3D~\cite{komorowski2021minkloc3d} offer efficiency, they suffer from quantization errors. Then, CASSPR~\cite{Xia_2023_ICCV} mitigates via a dual-branch cross-attention design. Coarse estimates are typically refined using registration algorithms like ICP~\cite{segal2009generalized} or autoencoders~\cite{elbaz20173d}. Recently, research has shifted toward cross-modal alignment, such as image-to-point cloud localization~\cite{li2025placerecognitionmeetmultiple, li2024vxp, xia2024uniloc}, enhancing cross-environment generalization. However, these methods cannot address localization scenarios where LiDAR sensors are unavailable as inputs.

\textbf{3D Vision and Language.} The integration of 3D vision and natural language has recently drawn increasing research interest. Prabhudesai \etal~\cite{prabhudesai2019embodied} implicitly link language to 3D visual feature representations to predict object-level 3D bounding boxes. Other methods~\cite{achlioptas2020referit3d, chen2020scanrefer, yuan2021instancerefer, feng2021free} perform referential grounding, locating the most relevant 3D objects within a scene from textual descriptions. However, these approaches primarily focus on indoor environments.
Text2Pos~\cite{kolmet2022text2pos} represents the first attempt at large-scale outdoor localization using natural language. It identifies coarse candidate locations and refines the pose estimation in a subsequent stage. Subsequently, various studies, such as \cite{wang2023text, shang2025vehiclesceneinteractiontextdriven3d, Liu_Huang_Li_Shen_Wang_2025, Feng_2025_ICCV, shang2026spatialocleveragingmultilevelspatial, shang2026riemanniansymplecticgeometryhierarchical}, have introduced novel methodologies to further refine the recognition and localization accuracy. Typically, MambaPlace~\cite{shang2025mambaplace} and CMMLoc~\cite{xu2025cmm} further push performance boundaries by integrating advanced sequence modeling techniques, namely the Mamba and Cauchy Mix modules, respectively.

\section{Problem statement}
\label{problem_statement}
We define a large-scale 3D reference map as
\[
M_{\mathrm{ref}}=\{\, s_i \mid i=1,\ldots,M \,\},
\]
where each cubic submap \(s_i\) is a set of 3D object instances
\[
s_i=\{\, p_i^j \mid j=1,\ldots,n_{s_i} \,\},
\]
and \(n_{s_i}\) denotes the number of instances in \(s_i\).
Let \(\{\, t_i \mid i=1,\ldots,M' \,\}\) be the collection of text descriptions;
each description \(t_i\) consists of spatial hints \(\{ h_i^k \}_{k=1}^{n_h}\),
where each \(h_i^k\) specifies a relation between the target location and an object instance. (M and M' is the number of 3D reference map and text descriptions)

Following~\cite{kolmet2022text2pos}, we adopt a two-stage \emph{coarse-to-fine} scheme.
In the coarse stage (text--submap global place recognition), a function \(F\) jointly encodes
a text description \(t\) and a submap \(s \in M_{\mathrm{ref}}\) into a shared embedding space,
such that matched text--submap pairs are pulled together while unmatched pairs are pushed apart.
In the fine stage, a matching-free network directly regresses the target position from the text
and the retrieved submaps. The identical text query $t$ is fed to both the coarse place recognition and the fine localization stages.

Accordingly, learning language-guided 3D localization reduces to predicting the ground-truth
planar coordinates \((x,y)\) (with respect to the scene frame) via
\begin{equation}
\label{eq:encoder}
\underset{\phi,\,F}{\min}\;
\mathbb{E}_{(x,y,t)\sim\mathcal{D}}
\Big\|
(x,y) - \phi\ \big(
t,\;
\underset{s \in M_{\mathrm{ref}}}{\operatorname*{argmin}}\;
d\ \big(F(t),F(s)\big)
\big)
\Big\|^{2},
\end{equation}
where \(d(\cdot,\cdot)\) is a distance metric (e.g., Euclidean), \(\mathcal{D}\) denotes the dataset,
and \(\phi\) maps a text description and a submap to fine-grained coordinates from a text description $t$ and a submap $s$.

\section{Methodology}
Fig.~\ref{fig:pipeline} shows our Text2Loc++ architecture. Global place recognition first matches the linguistic content within the point cloud map
to narrow down the search to a specific reference submap, while fine localization then
identifies the precise coordinates within that submap.
To accomplish text-to-submap retrieval, we employ separate text and submap branches to extract modality-specific features. A masked instance modality-aware contrastive learning framework is then used to train the model for precise multimodal alignment. Additionally, to cope with varying levels of textual complexity, we introduce a text distillation mechanism. These components, described in Section~\ref{section: coarse stage}, enable Text2Loc++ to adapt not only to varying levels of textual input complexity, but also to diverse urban environments and point cloud configurations, allowing it to retrieve a set of candidate submaps that potentially contain the target location.
Subsequently, the retrieved submaps are utilized to refine the estimated position via a specifically designed fine localization module, detailed in Section.~\ref{section: fine stage}.

\begin{figure*}[h]
  \centering
\includegraphics[width=1.0\linewidth]{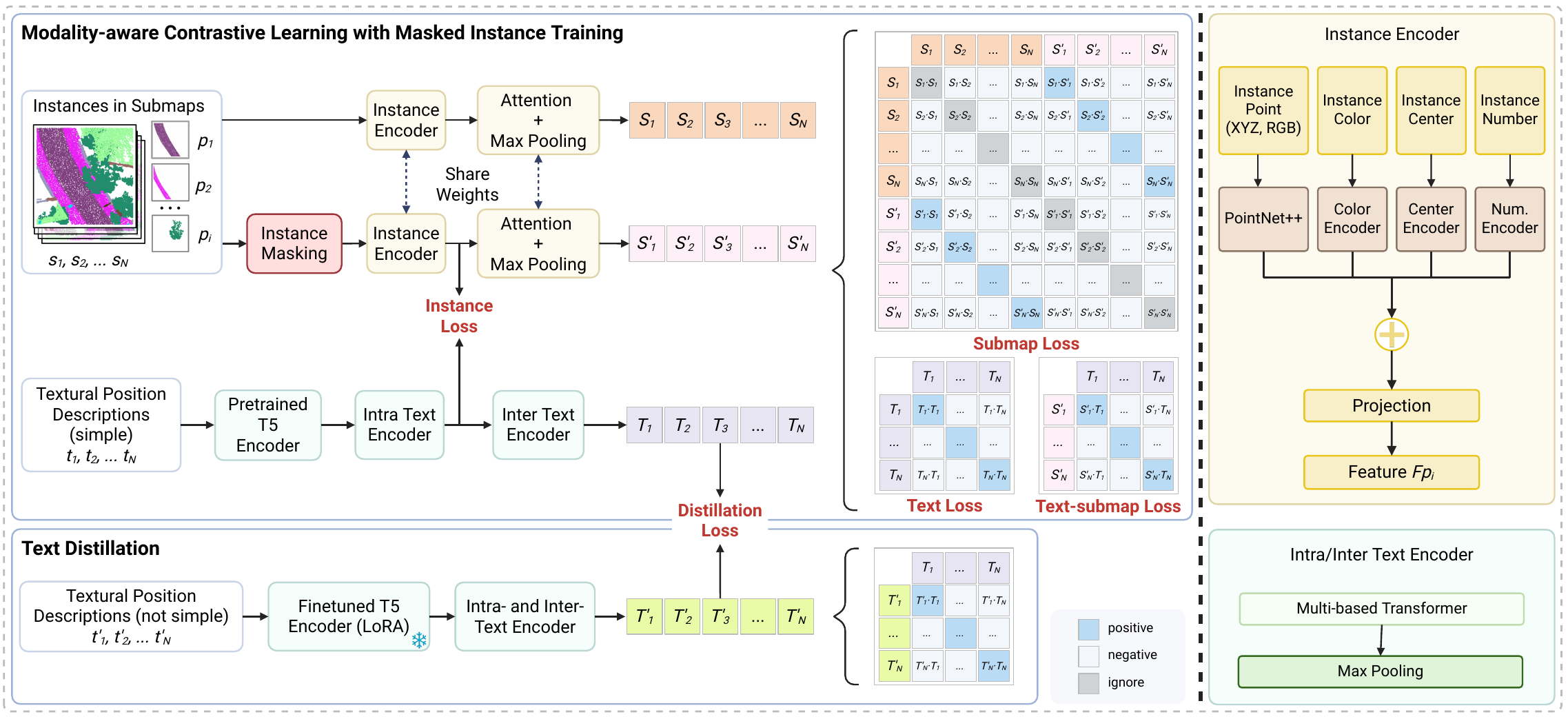}
  \caption{\textit{(left)} The architecture of global place recognition and the training procedure in the global place recognition. \textit{(right)} The architecture of instance encoder as well as intra/inter-text encoder architecture for point clouds. Note that the pre-trained/finetuned T5 model is frozen.}
  \label{fig:coarse}
  \vspace{-1em}
\end{figure*}

\begin{figure}[h]
    \centering
    \includegraphics[width=0.48\textwidth]{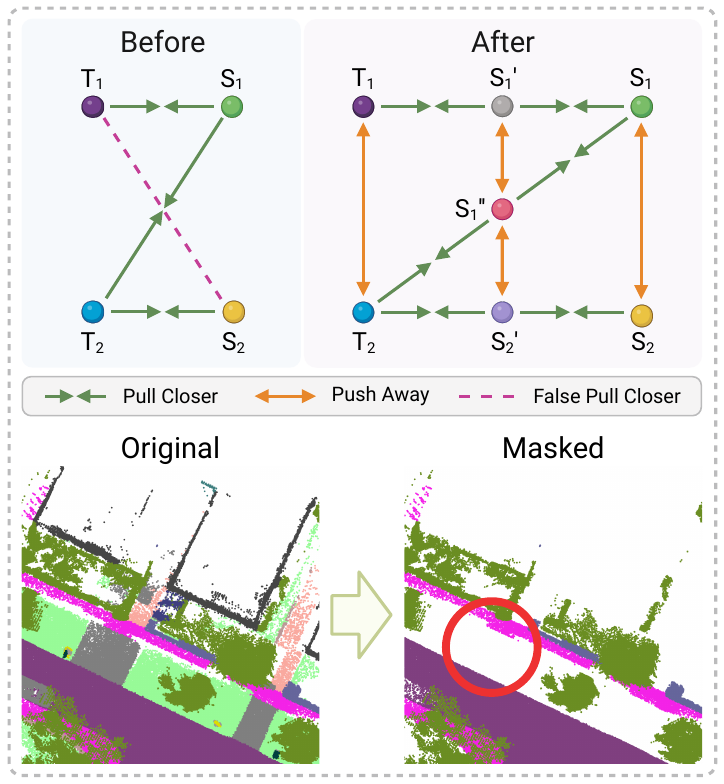}

    \caption{\textit{(top)} Comparison for without and with masked instance modality-aware contrastive learning in the training process. $T_1$ and $T_2$ are different text embeddings. $S_1$ and $S_2$ are different submap embeddings from $s_1$ and $s_2$. $S_1'$,  $S_1''$, and $S_2'$ are submap embeddings with masked instance process. 
    \textit{(bottom)} Visualization of original submap and masked submap. The red circle points out the masked instance. }
    \label{fig:coarse2}
    \vspace{-1em}
\end{figure}

\subsection{Global Place Recognition}
\label{section: coarse stage}

3D point cloud based place recognition is formulated as a retrieval problem, in which the task is to identify the most similar reference scan and its corresponding spatial location for a given query LiDAR scan. This is typically achieved by comparing the query’s global descriptor with those extracted from a database of reference scans, where the similarity is measured using a predefined descriptor distance metric.
Building on this paradigm, we extend the concept to a \textit{text–submap} cross-modal global place recognition framework for coarse localization. In this setting, the objective is to retrieve the submap that best corresponds to a natural language description. The key challenge lies in learning global descriptors for 3D submaps $S$ and textual queries $T$ that are both robust to modality-specific variations and discriminative enough to distinguish fine-grained spatial contexts. Following prior works~\cite{kolmet2022text2pos, wang2023text}, we employ a dual-branch encoding architecture that maps submaps $S$ and textual descriptions $T$ into a shared embedding space, as depicted in Fig.~\ref{fig:coarse} (left).

{\bf Text Branch.} We begin by utilizing a frozen, pre-trained large language model (T5)~\cite{2020t5} to extract fine-grained semantic representations from textual descriptions, thereby improving the quality of text embeddings. To further capture contextual dependencies, we design a hierarchical transformer architecture equipped with max-pooling layers (HTM). This module models both intra-sentence relationships through self-attention and inter-sentence semantics by aggregating information shared across sentences, as illustrated in Fig.~\ref{fig:coarse} (right bottom). For a given textual input, we first derive individual sentence embeddings for each spatial hint via the intra-text encoder. Subsequently, these embeddings are aggregated through the inter-text encoder to produce a global textual representation. The transformer serves as our intra- and inter-text encoder.
Additional architectural details are provided in the Supplementary Material.

{\bf 3D Submap Branch.} Each instance $P_i$ in the submap $S_N$ is represented as a point cloud that has both spatial and color (RGB) coordinates, forming the 6D features\footnote{The 6D-feature input is for the point cloud with color. If the point cloud is without color information, we only include x-y-z coordinates.} (Fig.~\ref{fig:coarse} (right top)).  
To extract semantic features from these points, we employ the most efficient 3D encoder - PointNet++~\cite{qi2017pointnet++}, though any more advanced encoder could be substituted. In parallel, we generate a color embedding by encoding RGB values using a color encoder and a positional embedding by encoding the instance centroid $\Bar{P_i}$ (i.e., the mean coordinates) through a positional encoder.
Since different object instances typically contain varying numbers of points(for example, road surfaces usually have more than 1,000 points, whereas poles have fewer than 500), we introduce a number encoder to explicitly represent this variation. Encoding the point count provides a useful class-specific prior that enhances instance differentiation.
All encoders (color, positional, and number) are implemented as three-layer multilayer perceptrons (MLPs), producing feature vectors with dimensions aligned to the semantic point embeddings. The resulting semantic, color, positional, and numerical embeddings are concatenated and passed through a projection MLP layer
that outputs the final instance embedding ${F}_{p_i}$. 
Finally, the set of \textit{in-submap} instance descriptors $\{{F}_{p_i}\}^{N_p}_{i=1}$  is aggregated into a global submap descriptor ${F}^S$ using an attention layer~\cite{xia2021soe} followed by a max pooling operation. 


{ \bf Masked Instance Training.}  Point clouds contain detailed and structured spatial information, whereas language is inherently unstructured and context-dependent. In general, it is impractical for people to describe every detail around them during localization. Therefore, we construct our text-description queries based on a representative subset of the surrounding scene. This modality and information gap introduces ambiguity, where a single submap may correspond to various textual descriptions, and a single text may relate to multiple submaps based on interpretation.
As shown in Fig.~\ref{fig:coarse2} (top left), $T_i=F(t_i),(i=1,2)$ and $S_i=F(s_i),(i=1,2)$ are text and submap embeddings, respectively. The definitions for $t_i$ and $s_i$ are in Section~\ref{problem_statement}. This example highlights the issue discussed above: the submap $s_1$ matches both the textual descriptions $t_1$ and $t_2$, while $t_2$ also corresponds to both $s_1$ and $s_2$. There is an example shown in Supplementary Material.
When applying standard contrastive learning, the model is incorrectly encouraged to pull $T_1$ closer to $S_2$, introducing semantic inconsistency (as $t_1$ doesn't describe the submap $s_2$). This illustrates a fundamental limitation of naive contrastive objectives in the presence of many-to-many cross-modal correspondences.

To address this issue, we propose a novel Masked Instance Training (MIT) strategy. In short, MIT identifies and randomly "masks out" the objects in the 3D submap that do not correspond to the text. Fig.~\ref{fig:coarse2} (bottom) illustrates a concrete example, where the bird's-eye view of the submap reveals that objects within the red circle in the masked submap have been masked out from the original submap.
Specifically, each submap is defined as ${ s_{i} = s_{i}^l \cup s_{i}^{nl}}$, where $s_{i}^l$ denotes the set of instances described in the text,  and $s_{i}^{nl}$ includes those that are not. 
During training, we randomly sample a subset $s_{i}^{'nl} \subseteq s_{i}^{nl}$ and construct a masked submap ${ s'_{i} = s_{i}^l \cup s_{i}^{'nl}}$. This encourages the model to focus on semantically aligned content while being robust to irrelevant or noisy instances.
A detailed algorithmic description is provided in the Supplementary Material. The proposed MIT improves the alignment of the cross-modal under many-to-many matching conditions, as validated by Ablation Studies. 

{\bf Modality-aware Contrastive Learning.} 
To address the "false pull" issue observed in Fig.~\ref{fig:coarse2} (top left), we introduce additional loss functions built upon the MIT baseline. As illustrated in Fig.~\ref{fig:coarse2} (top right), our proposed framework incorporates: a text-submap loss designed to better pull closer the embeddings of corresponding modalities; and a fine-grained instance loss to capture detailed object features. Conversely, to enhance discriminative power, we employ a text loss to push away embeddings with distinct meanings, along with a submap loss that separates original submaps from their masked counterparts. Together, these components constitute the Modality-aware Contrastive Learning (MCL) framework.

For significant cross-modal alignment, the input batch consists of 3D submap descriptors $\{S_i\}^{N} _{i=1}$ and matching text descriptors $\{T_i\}^{N}_{i=1}$ where $N$ is the batch size. We utilize the proposed MIT strategy to generate the 3D submap descriptors $\{S'_i\}^{N} _{i=1}$, which include only part of the instances in certain submaps. 
\noindent
We define that $\tau$ is the temperature coefficient, similar to CLIP~\cite{radford2021learning}. Within a training mini-batch, the text-submap loss can be described as $L_{\text{align}}$:
\begin{equation}\footnotesize
       L_{\text{align}} = \frac{1}{N} \sum_{i\in N} \left [ -\log\frac{\exp(T_i\cdot S'_{i}/\tau)}{\sum\limits_{j\in N} \exp(T_i\cdot S'_{j}/\tau)} - \log\frac{\exp(S'_i\cdot T_{i}/\tau)}{\sum\limits_{j\in N} \exp(S'_i\cdot T_{j}/\tau)} \right ].
\end{equation}

As discussed in the last section, text-to-submap matching inherently exhibits a many-to-many correspondence, making it difficult to establish precise alignments between textual descriptions and point cloud data; consequently, constraining solely at the submap level proves inadequate for capturing fine-grained cross-modal relationships. Thus, we incorporate an instance-level loss to constrain the alignment between sentence embeddings and the corresponding instance embeddings within the submap. Given an input batch of 3D instance descriptors  $\{{I}^S_i\}^{N_I}_{i=1}$ and matching text descriptors $\{{I}^T_i\}^{N_I}_{i=1}$ with the instance where $M$ is the number of the number of the instance. For simplicity, we take a fixed number of instances in each batch $N_I=kN (k \in \mathbb{N^{+}})$. The instance loss $L_{\text{inst}}$ is

\begin{equation}\footnotesize
       L_{\text{inst}} = \frac{1}{N_I} \sum_{i\in M} \left [ -\log\frac{\exp({I}^T_i\cdot {I}^S_i/\tau)}{\sum\limits_{j\in M} \exp({I}^T_i\cdot {I}^S_j/\tau)} - \log\frac{\exp({I}^S_i\cdot {I}^T_i\tau)}{\sum\limits_{j\in M} \exp({I}^S_i\cdot{I}^T_j/\tau)}\right ].
\end{equation}

To address the limitations illustrated in Fig.~\ref{fig:coarse2}, we aim to construct a semantically meaningful embedding space that requires not only cross-modal supervision but also modality-aware constraints.
For text embeddings, it is essential that distinct descriptions are mapped to well-separated regions in the embedding space. To enforce this, we introduce a text loss $L_{\text{text}}$ as that explicitly encourages inter-text separation.

\begin{equation}\footnotesize
    L_{\text{text}} =\frac{2}{N} \sum_{i\in N} -\log\frac{\exp(T_i\cdot T_i/\tau)}{\sum\limits_{j\in N} \exp(T_i\cdot T_{j}/\tau)} . 
\end{equation}

We approach the representation of submap embeddings from two primary perspectives: (1) ensure sufficient separation between embeddings of different submaps, and (2) maintain consistency between each submap and its masked variants, which share the same semantic description and should thus remain close despite perturbations~\cite{chen2020simpleframeworkcontrastivelearning}.
To jointly enforce these objectives, we propose a double contrastive learning framework as submap loss ($L_{\text{submap}}$) that penalizes mismatched submap–masked submap pairs while promoting alignment across multiple representations of the same semantic content.

{\footnotesize
\begin{align}
L_{\text{submap}} = \frac{1}{N} \sum_{i=1}^{N} \biggl[ 
& -\log \frac{\exp(S_i \cdot S'_i / \tau)}{\sum_{j=1}^{N} \exp(S_i \cdot S'_j / \tau) + \sum_{j \neq i}^{N} \exp(S'_i \cdot S'_j / \tau)} \nonumber \\
& -\log \frac{\exp(S'_i \cdot S_i / \tau)}{\sum_{j=1}^{N} \exp(S'_i \cdot S_j / \tau) + \sum_{j \neq i}^{N} \exp(S_i \cdot S_j / \tau)} \biggr] .
\end{align}
}

After applying masked instance modality-aware learning, the training process is illustrated in Fig.~\ref{fig:coarse2} (top right). In conclusion, the final loss formulation is:
\begin{equation}\small
       L = \alpha_1 L_{\text{align}} + \alpha_2 L_{\text{inst}} +  \alpha_3 L_{\text{submap}} + \alpha_4 L_{\text{text}} .
\end{equation}

{ \bf Text Distillation.} As textual descriptions increase in complexity, the modality gap between language and point cloud data, along with the model’s limited capacity to simultaneously extract fine-grained semantics and perform accurate cross-modal alignment, renders direct matching ineffective. To mitigate this challenge, we propose a stepwise training strategy based on text distillation, which progressively guides the model toward robust understanding and alignment of complex linguistic inputs with 3D spatial representations First, we perform multimodal contrastive learning between simple descriptions and point clouds. Then, we freeze the previously trained network and introduce a new text module to handle moderate or complex text descriptions, generating new text embeddings. Since the original pretrained T5 module is no longer effective in processing complex text, we finetune the T5 encoder using the LoRA approach \cite{Hu2021LoRALA}. The detail finetune steps are in the Supplementary Material. For content-equivalent simple text descriptions, we use the original frozen text module to produce the old text embeddings. In the embedding space, these two representations are expected to be close due to their underlying semantic consistency. Consequently, we employ a distillation loss, formulated as a contrastive learning objective, to align them. A more thorough comparative study is in Section ~\ref{sec:multiple_analysis}.

\begin{figure}[t!]
  \centering
  \includegraphics[width=1.0\linewidth]{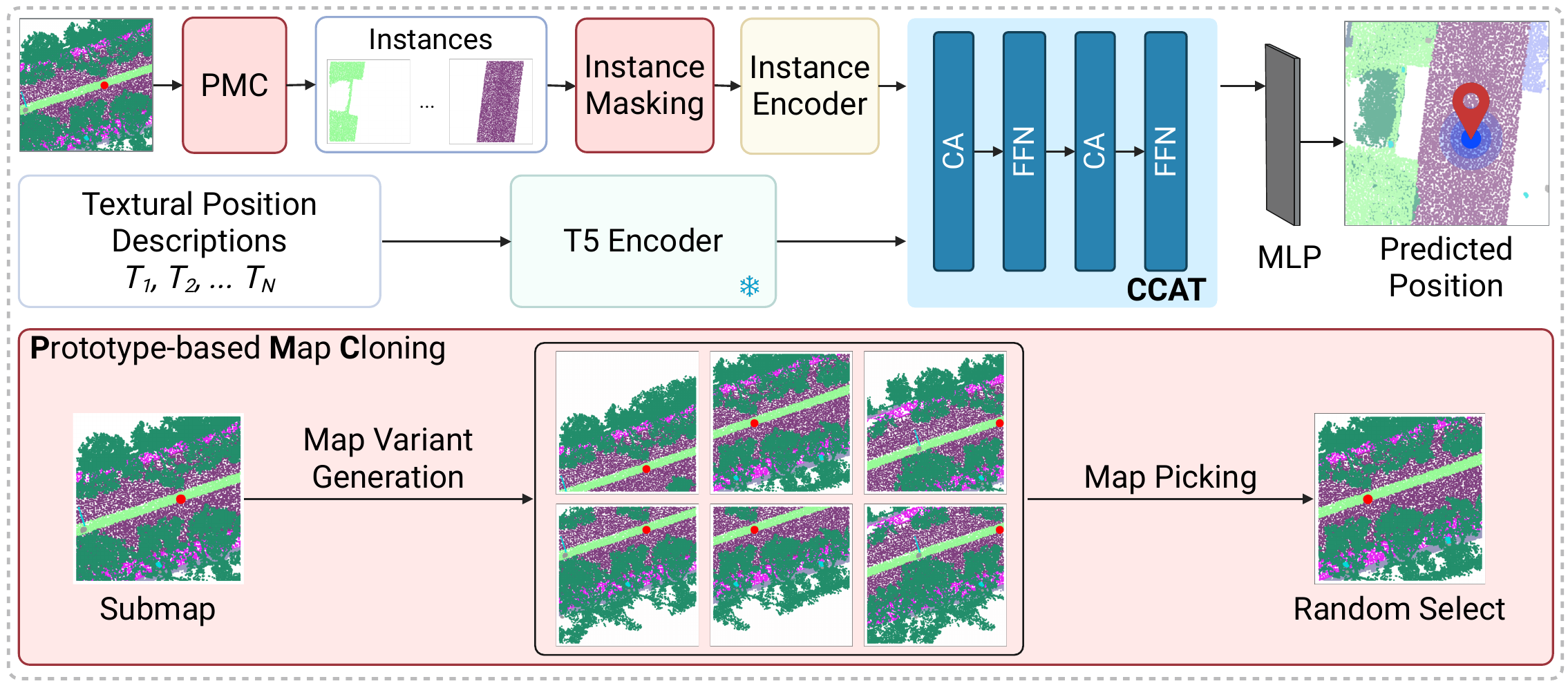}
  \caption{\textit{(top)} The proposed matching-free fine localization architecture. It consists of two parallel branches: instance encoder as the top branch and text encoder as the bottom branch. Cascaded Cross-Attention Transformers (CCAT) fuse the multimodal information. \textit{(bottom)} The procedure of the Prototype-based Map Cloning (PMC).}
  \label{fig:fine-loc}
  \vspace{-1em}
\end{figure}

\subsection{Fine Localization}
\label{section: fine stage}

Following the text–submap global place recognition stage, our objective in fine localization is to refine the predicted target position within the retrieved submaps. The network is then supervised to predict a Bird-eye-view 2D $(x, y)$
coordinate, which represents the relative position of the location within the submap,
with values normalized between $[0, 1]$. We utilize ground-truth coordinates as the
supervision signal. 
While prior approaches~\cite{kolmet2022text2pos, wang2023text} have achieved promising results using text–submap matching strategies, the inherent ambiguity of natural language descriptions often leads to unreliable offset predictions for individual object instances. To overcome this limitation, we introduce a matching-free fine localization network, as illustrated in Fig.~\ref{fig:fine-loc}.

In this framework (Fig.~\ref{fig:fine-loc} top), the text branch (bottom) extracts fine-grained linguistic features using a frozen or fine-tuned pre-trained language model, T5~\cite{2020t5}, followed by an attention module and a max-pooling operation. The submap branch (top) employs a prototype-based map cloning (PMC) module to enrich the diversity of submap variants and then extracts point cloud features using the same instance encoder utilized in the global place recognition stage. 
Similarly, we also introduce the MIT strategy, which randomly masks a subset of irrelevant 3D instances within the submap during training, forcing the fusion network to focus on effective spatial signals while ignoring the interference of environmental noise and irrelevant features. Subsequently, the text and submap features are integrated through a Cascaded Cross-Attention Transformer (CCAT), after which a lightweight multi layer perceptron (MLP) regresses the final target position.


{\bf Cascaded Cross-attention Transformer (CCAT).} To effectively model the interactions between the text and 3D submap branches, we introduce a Cascaded Cross-Attention Transformer (CCAT) for multimodal feature fusion. Following the design of~\cite{Xia_2023_ICCV}, the CCAT is composed of two sequential Cross-Attention Transformer (CAT) modules, each contains a cross attention module (CA) and a feed-forward network (FFN).

In the first CA, the point cloud features serve as the Query, while the text features act as the Key and Value. This allows the model to refine the point representations by incorporating semantic cues from the text, yielding text-informed point feature maps. Conversely, in the second CA, the text features are used as the Query, and the enhanced point features from the first CAT are used as the Key and Value, resulting in text features enriched with geometric context.

Unlike the hierarchical structure proposed in~\cite{Xia_2023_ICCV}, our design employs a cascaded configuration of the first and the second CAT, enabling progressive cross-modal interaction. In this work, we utilize two stacked CCAT blocks to achieve more comprehensive feature fusion. Further ablation studies and analyses are provided in the Supplementary Material.

{\bf Prototype-based Map Cloning (PMC).} To generate more diverse and informative submap variants for training, we introduce a Prototype-based Map Cloning (PMC) module in the training time. For each pair $\{t_i, s_i\}$, our goal is to construct a collection of neighboring submap variants $ \mathcal{G}_i$ centered around the current map $S_i$, which can be formulated as follows: 

\begin{equation}
\begin{aligned}
    \mathcal{G}_i = \{S_j \; | \; & \big \| {\Bar{s_j} - \Bar{s_i}} \big \|_{\infty}< \alpha ,
    \; \big \|\Bar{s_j} - c_i \big \|_{\infty} < \beta \; \},
\end{aligned}
\end{equation}
\noindent
where $\Bar{s_i}$, $\Bar{s_j}$ are the center coordinates of the submaps $s_i$ and $s_j$ respectively.
$c_i$ represents the ground-truth target position described by $t_i$, $\alpha$ and $\beta$ are the pre-defined thresholds.

In practice, we observe that certain submaps in $ \mathcal{G}_i$ don't have enough object instances corresponding to the textual descriptions $T_i$. To mitigate this issue, we apply a filtering procedure by enforcing a minimum instance threshold, allowing at most $N_m$ instance mismatch. After this filtering step, a single submap is randomly sampled from the refined $ \mathcal{G}_i$ for training.

{ \bf Loss function.} In contrast to previous approaches~\cite{kolmet2022text2pos, wang2023text}, our fine localization network eliminates the text–instance matching module, resulting in a simpler and more efficient training process. It is important to note that this network is trained independently from the global place recognition stage. The objective of training is to minimize the distance between the predicted target position and the ground truth. To this end, we employ a mean squared error loss $L_{r}$ as the sole optimization objective for the translation regressor.
\begin{equation} 
\begin{aligned}
L(C_{gt}, {C}_{pred}) = \big \|C_{gt} - {C}_{pred}   \big \|_{2},
\end{aligned}
\label{Eq: regression loss}
\end{equation}
\noindent 
where $C_{pred}=(x, y)$ (see Eq. (\ref{eq:encoder})) is the predicted target coordinates, and ${C}_{gt} $ is the ground-truth coordinates.

\section{Experiments}
\subsection{Experimental Datasets}

The KITTI-360 dataset offers a large number of scenes with notable geographic and point cloud variations between training and testing sets. However, all data are collected in the same city using the same LiDAR device, and each pose is precisely georegistered in OpenMap, resulting in a high level of consistency. Therefore, to evaluate a model’s generalization capability, it must maintain stable performance across diverse cities, devices, and acquisition settings. To this end, we incorporate more varied data sources to evaluate robustness and generalization.

We train and evaluate the proposed Text2Loc++ on six benchmark datasets designed for text-based point cloud localization. This includes the two KITTI360Pose benchmarks \cite{kolmet2022text2pos} and four additional datasets containing point clouds with semantic annotations: Toronto ~\cite{tan2020toronto3d}, Paris\_CARLA~\cite{deschaud2021pariscarla3d}, TUM~\cite{zhu2020tum}, and Paris\_Lille~\cite{roynard2017parislille3d}.
The two KITTI360Pose datasets share identical submap structures and textual descriptions; the only distinction lies in the presence or absence of color information in the point clouds. Specifically, the Toronto and Paris\_CARLA datasets provide colored point clouds, whereas the TUM and Paris\_Lille datasets contain only geometric data and semantic annotations. With the exception of the KITTI360Pose dataset, we augment all point clouds with textual descriptions according to the protocol proposed in \cite{kolmet2022text2pos}. Additionally, we standardize semantic labels across all datasets to ensure consistency in representation across different urban environments.



\begin{table*}[h]
	\centering\small
	\begin{tabular}{cccccccccc}
        \toprule
		 &  & \multicolumn{4}{c}{Baseline} & \multicolumn{4}{c}{Refinement} \\
		\cmidrule{3-10}
        Dataset & Color & \multicolumn{2}{c}{Training} & \multicolumn{2}{c}{Testing} & \multicolumn{2}{c}{Training} & \multicolumn{2}{c}{Testing} \\
		\cmidrule{3-10}
        & & \# submaps & \# texts & \# submaps & \# texts & \# submaps & \# texts & \# submaps & \# texts \\
        \midrule
Kitti360 & \cmark  & 11259 & 28807 & 4308 & 11505 & 11259 & 28807 & 4308 & 11505 \\
Paris\_CARLA & \cmark & - & - & 1004 & 3857  & 2936 & 10855 & 1004 & 3857 \\
Toronto & \cmark & - & - & 685 & 2087  & - & - & 685 & 2087 \\
        \midrule
Kitti360 (no color) & \xmark  & 11259 & 28807 & 4308 & 11505 & 11259 & 28807 & 4308 & 11505 \\
Paris\_Lille   & \xmark & - & - & 185 & 845 & 566 & 2337 & 185 & 845 \\
TUM     & \xmark & - & - & 183 & 755 & 350 & 1518 & 183 & 755 \\
        \bottomrule
	\end{tabular}
	\caption{Dataset information. Number of training and testing submaps and texts for our baseline and refinement networks. Color means whether the point cloud contains color information.}
	\label{tab:dataset_split}
\end{table*}

\begin{figure}[t]
  \centering
  \setlength{\tabcolsep}{2pt} 
  \begin{tabular}{c: c}
    \includegraphics[width=0.55\linewidth]{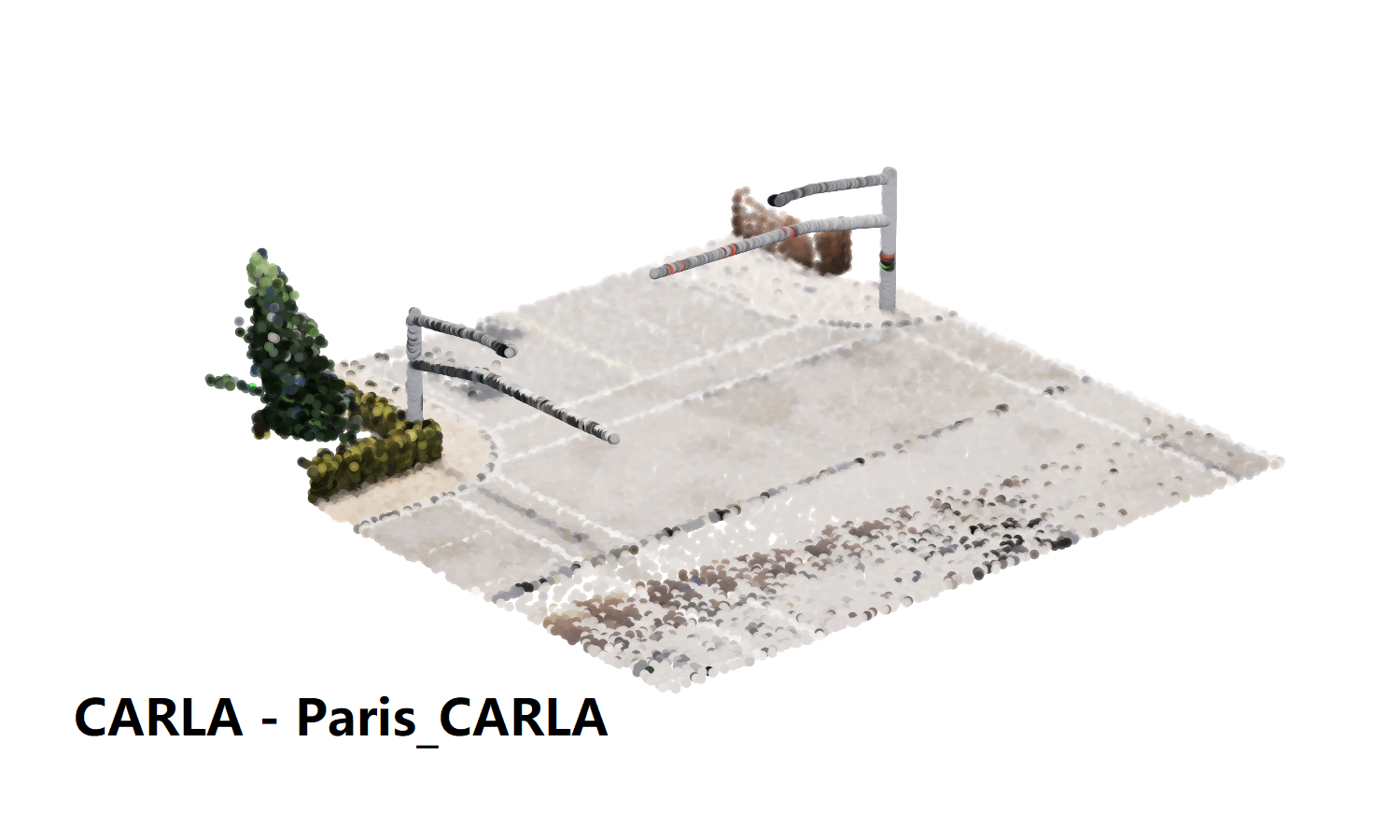} &
    \includegraphics[width=0.45\linewidth]{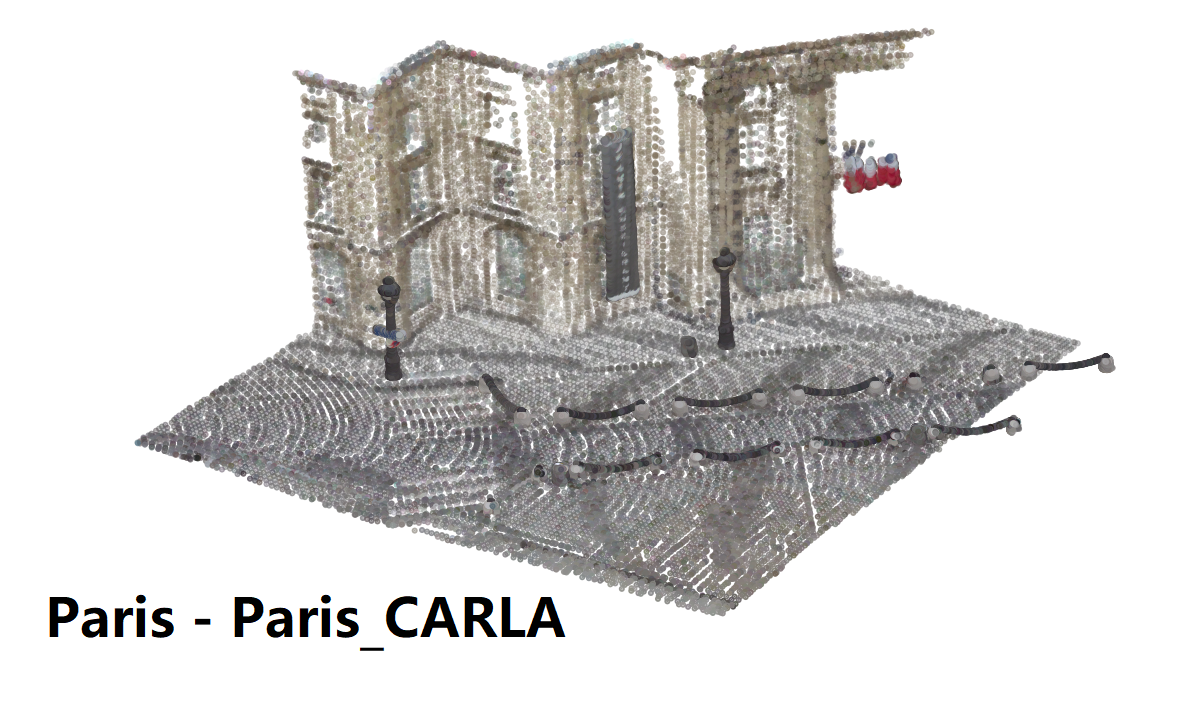} \\
    \hdashline
    \includegraphics[width=0.45\linewidth]{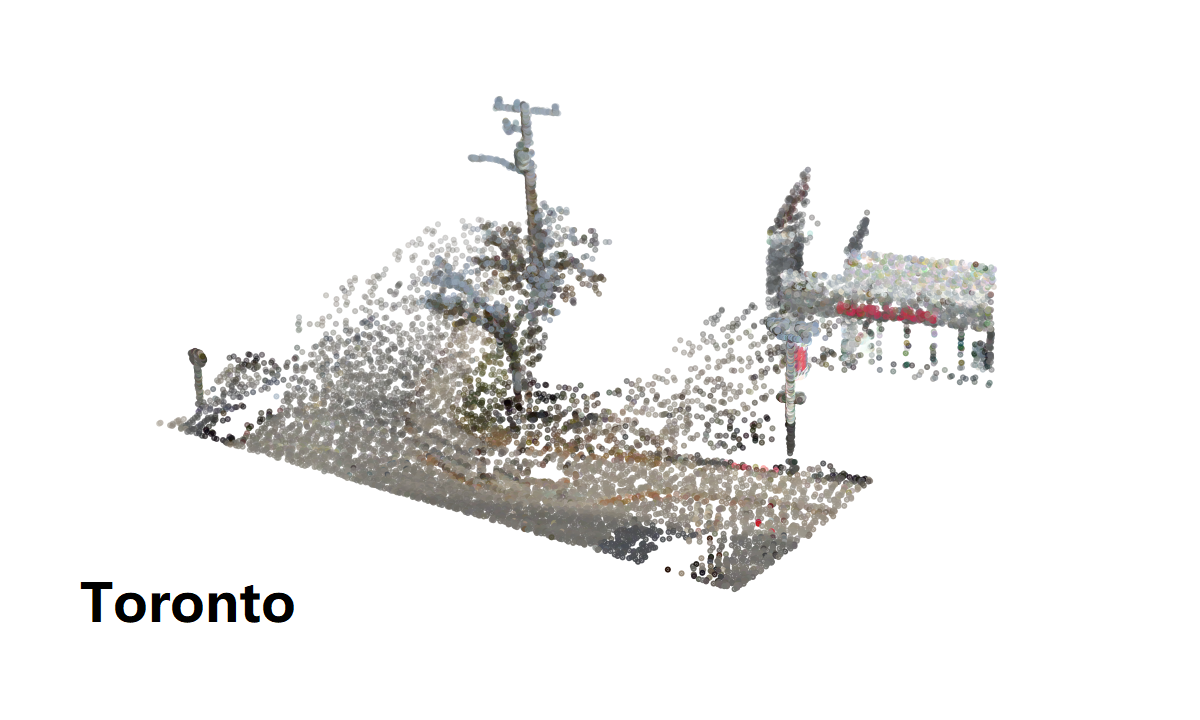} &
    \includegraphics[width=0.45\linewidth]{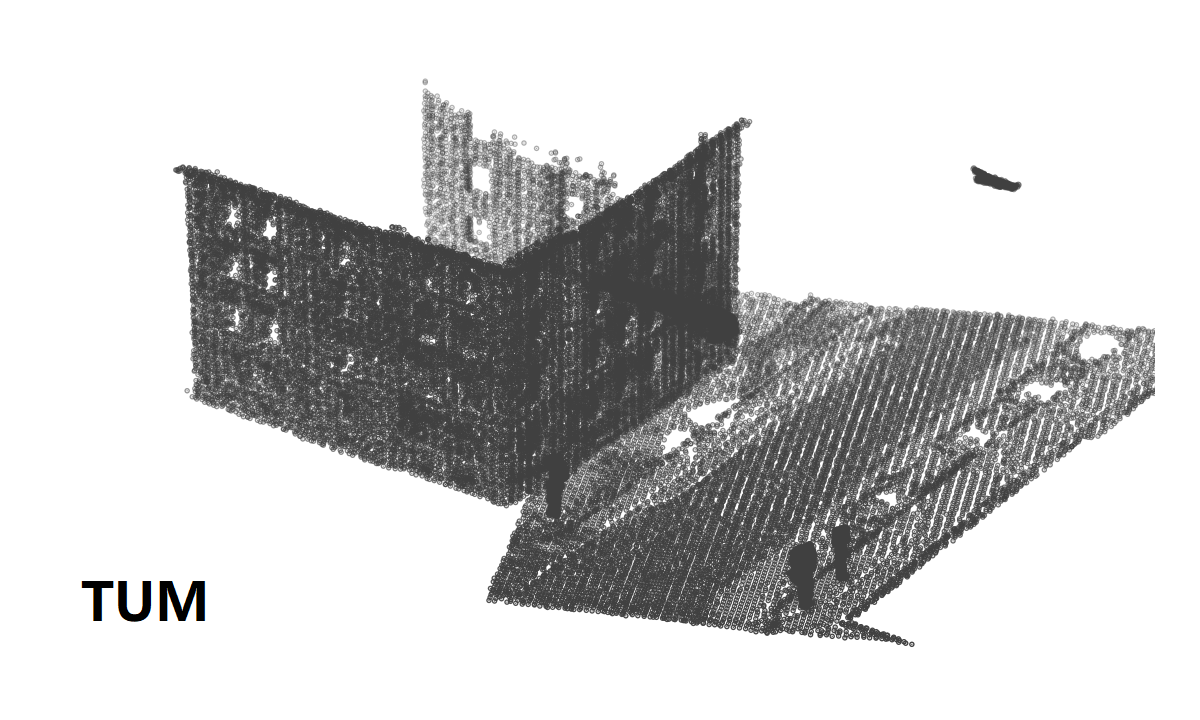} \\
    \hdashline
    \includegraphics[width=0.45\linewidth]{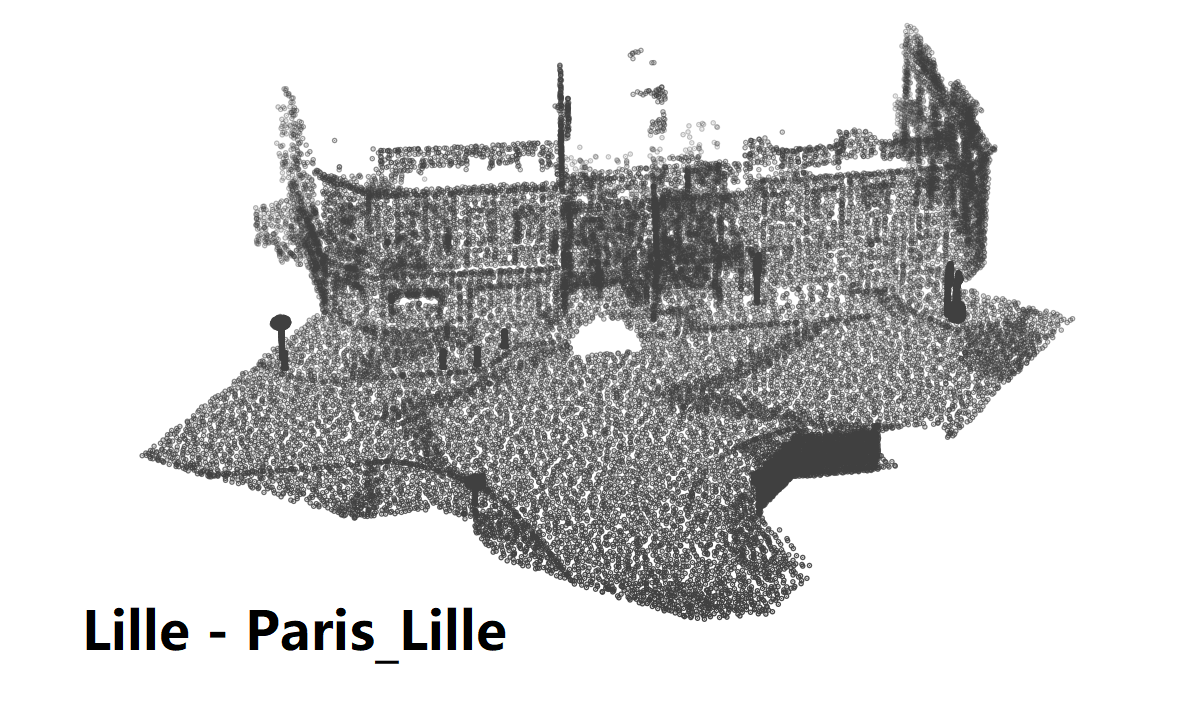} &
    \includegraphics[width=0.45\linewidth]{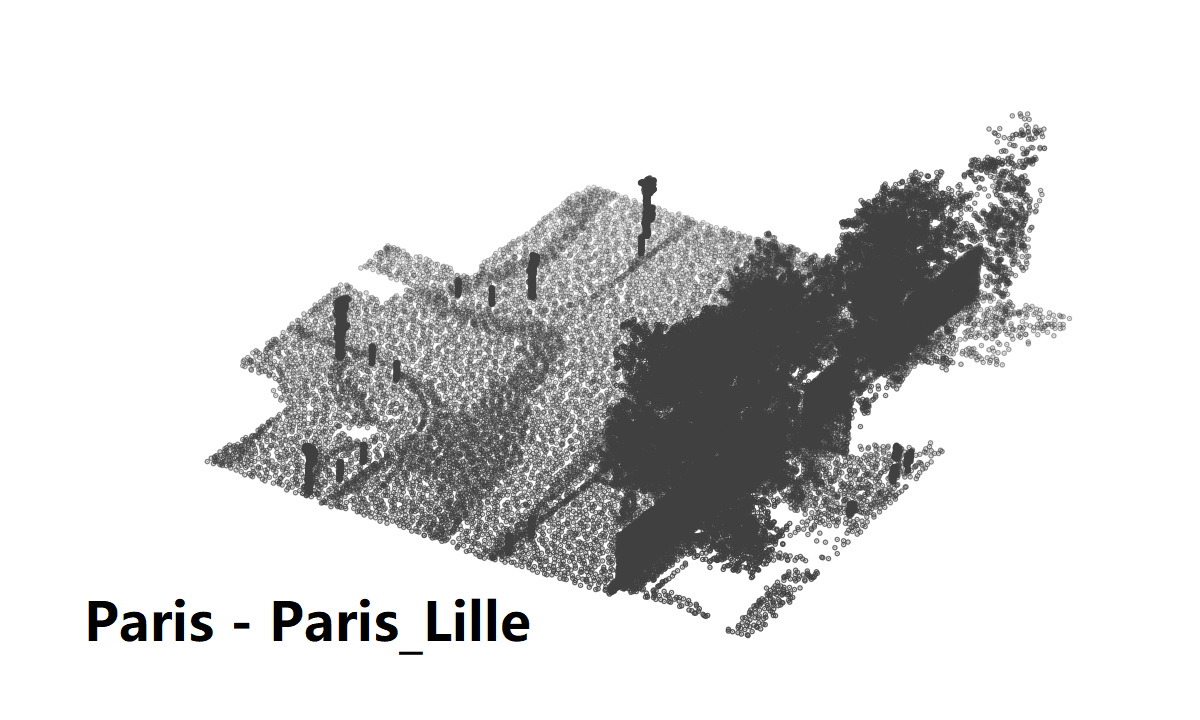}
  \end{tabular}
  \caption{\textbf{Samples of point cloud from different datasets.} We present visualizations of the point clouds from the four constructed benchmark datasets. Subfigures CARLA - Paris\_CARLA, Paris - Paris\_CARLA, and Toronto depict datasets with color information, whereas other subfigures correspond to datasets without color attributes.}
  \label{fig:dataset_point}
\end{figure}

\subsubsection{Data Specification}
Except from Kitti360, the other datasets offer annotated point clouds collected from urban environments with different schemes of semantic categories. We visualize the new submaps generated by ourselves in Fig.~\ref{fig:dataset_point}.  For Toronto and TUM datasets, they only have 8 raw semantic categories. Thus, we follow the semantic class definition in KITTI360~\cite{Liao2021ARXIV} to split and annotate the point clouds. Since we only focus on the layout of the environment, the dynamic objects are removed from the raw point clouds. For Paris\_CARLA and Paris\_Lille datasets, we directly convert the label if the name of the category exists in the KITTI360 class definition. If not, we utilize the class definition in~\cite{Cordts2016Cityscapes}. Then, the submaps and texts are created in the same way of KITTI360Pose. In detail, the 3D submap is a cube that is 30m long with a stride of 10m. Below are the detailed explanations for each dataset:

{\bf KITTI360Pose Dataset.}
We use the dataset constructed by \cite{kolmet2022text2pos}, which consists of point clouds from 9 districts, covering 43,381 position-query pairs over a total area of 15.51~$km^2$. Following \cite{kolmet2022text2pos}, we select five scenes (11.59~$km^2$) for training, one for validation, and the remaining three scenes (2.14~$km^2$) for testing. This results in 11,259 training submaps and 4,308 testing submaps, with a total of 15,567 submaps across the entire dataset.

{\bf Paris\_CARLA Dataset} is a collection of dense, colored point clouds captured in outdoor environments. The dataset consists of two parts: a synthetic set generated with the open-source CARLA simulator (\SI{700}{M} points), and a real-world set collected in central Paris (\SI{60}{M} points). The synthetic portion covers approximately \SI{5.8}{km} of driving distance, while the real-world data spans \SI{550}{m} across three streets in Paris. The real point cloud is collected by a mobile platform equipped with a tilted Velodyne HDL-32 LiDAR (45° to the horizon) and a 360° Ladybug5 camera system (comprising 6 lenses). All data are annotated with 23 semantic classes and instance-level labels, consistent with the class definitions used in CARLA to ensure label alignment between synthetic and real domains.

{\bf Toronto Dataset.}
Toronto is a large-scale outdoor urban point cloud dataset collected in Toronto, Canada, using a vehicle-mounted mobile LiDAR scanning (MLS) system. The platform is equipped with a 32-line LiDAR sensor, a Ladybug5 panoramic camera, a GNSS system, and a SLAM module. Natural RGB colors are assigned to each point using camera projections. The dataset spans approximately \SI{1}{km} of road and contains 78.3 million points, with UTM coordinates centered at (43.726, –79.417). All points are labeled with one of eight semantic classes.

{\bf TUM Dataset} is a mobile laser-scanning dataset covering approximately 80,000 $m^{2}$ with detailed semantic annotations. Collected by the Fraunhofer Institute of Optronics, System Technologies, and Image Exploitation (IOSB) using two Velodyne HDL-64E laser scanners, the dataset was later annotated by the Chair of Photogrammetry and Remote Sensing at TUM. It encompasses an urban area with around \SI{1}{km} of roadways and includes over 40 million annotated points, categorized into eight object classes. Notably, this dataset does not contain any images.

{\bf Paris\_Lille Dataset}  is a mobile LiDAR dataset collected using a Velodyne HDL-32E sensor. The dataset includes point clouds from two cities (Paris and Lille), covering nearly \SI{2}{km} and totaling over 140 million points. All points are annotated with fine-grained semantic labels across 50 classes.

\begin{table}[h]
\centering  
\scalebox{0.95}{
\begin{tabular}{l l}
\toprule
\multirow{6}{*}{\centering Simple}  & The pose is on-top of a green vegetation. \\ 
 & The pose is east of a dark-green fence. \\
 & The pose is north of a green sidewalk. \\
 & The pose is south of a green building. \\
 & The pose is north of a green terrain.   \\ 
 & The pose is north of a green vegetation. \\
\midrule
Moderate 
& \multicolumn{1}{m{7.0cm}}{Two vegetations, the sidewalk, the building and the terrain are green. One vegetation is below the pose. The sidewalk, the terrain and one vegetation are south of the pose. The building is north of the pose. They are east of the dark-green fence.} \\
\midrule
Complex
& \multicolumn{1}{m{7.0cm}}{In the landscape, two areas of vegetation, the sidewalk, the building, and the surrounding terrain display a vibrant green hue, contributing to the overall vitality of the scene. One patch of greenery is situated directly below the post, while the sidewalk, the terrain, and another clump of vegetation stretch out to the south of it. Meanwhile, the building stands to the north of the post, commanding attention within the layout. All these elements are positioned east of a dark-green fence, which serves as a contrasting boundary, framing the lively environment.} \\
\bottomrule
\end{tabular}
}
\vspace{1mm}
\caption{Textual descriptions of the same location in the submap at different levels of linguistic complexity.}
 \vspace{-1em}
\label{tab:text_mode_demo}  
\end{table}

\subsubsection{{Dataset Splitting}}
\label{dataset:data_splitting}
We adopt dataset-specific splitting strategies tailored to the structural characteristics of each benchmark. For datasets containing multiple traversals of the same region (e.g., KITTI360Pose, CARLA in Paris\_CARLA, and Paris\_Lille), training and testing reference maps are constructed from disjoint runs to ensure route-level separation. For datasets comprising only a single traversal, alternative protocols are employed: the entire Toronto dataset is reserved for evaluation, whereas in Paris in Paris\_CARLA and TUM, geographically distinct regions are partitioned into training and testing subsets.

Each reference map is subsequently partitioned into submaps at fixed spatial intervals along the trajectory to facilitate fine-grained correspondence. Additional implementation details concerning the segmentation procedure are provided in the Supplementary Material.

To evaluate cross-dataset generalization, we define two training paradigms: a baseline model, trained solely on the KITTI360Pose dataset, and a refinement model, trained jointly on KITTI360Pose and supplementary datasets.
Table~\ref{tab:dataset_split} summarizes the number of training and testing submaps used for the baseline and refinement networks.


\subsubsection{Textual Descriptions}
In real‐world communication, humans naturally describe self-locations by referencing multiple objects, their spatial relationships, and relevant attributes. In the prior publications, a text-to-point cloud system trained solely on minimal statements (e.g., 'the pose is on top of a gray road.'). This may struggle to capture the richness and nuance of human descriptions. By incorporating more complex narratives, where multiple elements and their relationships are mentioned, the model can develop a deeper, more human-like understanding of locations.
It learns to handle a wide range of linguistic expressions, from concise descriptions to detailed explanations (potentially containing redundant information), enabling it to process a broader spectrum of user queries. 
This not only improves the alignment between text and 3D geometry but also supports more natural and conversational interactions.
To systematically study the effect of text complexity on localization accuracy, we thus categorize text descriptions into three levels of complexity:

{\bf Simple}: Each sentence describes only the spatial relationship between the self-localized target and a single reference instance, with no inter-sentence connections. 

{\bf Moderate}: Building on the simple level, additional attributes such as direction, color, and instance labels are integrated, resulting in more concise and human-like descriptions. Moderate inter-sentence correlations are introduced. 

{\bf Complex}: Extending the moderate level, sentences are further diversified to resemble real-world text or speech-to-text outputs generated by large language models (LLMs)~\cite{openai2024gpt4technicalreport}. In this setting, descriptions may contain missing details such as omitted colors or instance labels, making the text more natural but also noisier. 

Table. \ref{tab:text_mode_demo} shows the distinct characteristics across different complexity levels in one color submap. The simple, moderate, and complex text descriptions each describe
the same spatial relationships between the target and instances within the submap, differing only in their level of linguistic
complexity. In other words, the three complexity levels convey the same amount of information about the same location. Increasing the descriptive complexity only makes the text appear richer in form, without adding any new spatial information. More detailed examples (including color and no-color submaps) and generation procedures are provided in the Supplementary Material.

\begin{table*}[h!]
    \centering
    \footnotesize
    \fboxsep0.75pt
    \setlength{\extrarowheight}{0.2pt}
    \begin{tabular}{p{1.2cm} p{2.4cm} >{\centering\arraybackslash}m{1cm} >{\centering\arraybackslash}m{1cm} >{\centering\arraybackslash}m{1cm} >{\centering\arraybackslash}m{1cm} >{\centering\arraybackslash}m{1cm} >{\centering\arraybackslash}m{1cm} >{\centering\arraybackslash}m{1cm} >{\centering\arraybackslash}m{1cm} >{\centering\arraybackslash}m{1cm}}
    \toprule
     &  & \multicolumn{9}{c}{Submap Retrieval Recall (\%) $\uparrow$} \\
    \cmidrule(lr){3-11}
     Setting & Method & \multicolumn{3}{c}{\textit{Simple}} & \multicolumn{3}{c}{\textit{Moderate}} & \multicolumn{3}{c}{\textit{Complex}} \\
    \cmidrule(lr){3-11}
     &  &  k=1 & k=3 & k=5  & k=1 & k=3 & k=5  & k=1 & k=3 & k=5 \\
    \midrule
    \multicolumn{11}{c}{KITTI360Pose~\cite{kolmet2022text2pos}} \\
    \midrule
     & Text2Pos~\cite{kolmet2022text2pos}  & 11.8 & 24.6 & 32.5 & 8.6 & 18.1 & 24.0 & 3.3 & 7.8 & 11.0 \\
     & Text2Loc~\cite{xia2024text2loc}     & 29.3 & 47.7 & 56.9 & 21.6 & 39.3 & 48.2 & 17.0 & 32.0 & 40.1 \\
Baseline
     & MambaPlace~\cite{shang2025mambaplace}     & 31.0 & 52.0 & 62.0 & - & - & - & - & - & -  \\
     & CMMLoc~\cite{xu2025cmm}             & 32.3 & 53.4 & 63.1 & - & - & - & - & - & -  \\
     & Text2Loc++ (Ours)                   & \overlaybold{\textbf{35.3}} & \overlaybold{\textbf{57.6}} & \overlaybold{\textbf{66.9}} & \overlaybold{\textbf{34.5}} & \overlaybold{\textbf{56.6}} & \overlaybold{\textbf{65.9}} & \overlaybold{\textbf{27.9}} & \overlaybold{\textbf{48.1}} & \overlaybold{\textbf{57.7}} \\
     \hdashline
     & Text2Pos~\cite{kolmet2022text2pos}  & 10.5 & 21.8 & 28.7 & 8.0 & 17.5 & 23.6 & 3.3 & 8.0 & 11.4 \\
Refinement     
     & Text2Loc~\cite{xia2024text2loc}     & 29.0 & 48.8 & 58.0 & 21.3 & 39.7 & 48.6 & 18.0 & 34.2 & 42.4 \\
     & Text2Loc++ (Ours)                   & \overlaybold{\textbf{34.6}} & \overlaybold{\textbf{56.5}} & \overlaybold{\textbf{65.6}} & \overlaybold{\textbf{34.0}} & \overlaybold{\textbf{56.8}} & \overlaybold{\textbf{65.4}} & \overlaybold{\textbf{26.2}} & \overlaybold{\textbf{46.0}} & \overlaybold{\textbf{55.7}} \\
    \midrule
    \multicolumn{11}{c}{Paris\_CARLA~\cite{deschaud2021pariscarla3d}} \\
    \midrule
     & Text2Pos~\cite{kolmet2022text2pos}  & 2.9 & 8.0 & 12.0 & 1.3 & 3.4 & 5.1 & 0.6 & 1.9 & 3.2 \\
     & Text2Loc~\cite{xia2024text2loc}     & 11.2 & 21.2 & 28.5 & 6.0 & 14.1 & 20.1 & 2.5 & 5.4 & 8.6 \\
Baseline
     & MambaPlace~\cite{shang2025mambaplace}    & 9.9 & 21.4 & 29.1 & - & - & - & - & - & - \\
     & CMMLoc~\cite{xu2025cmm}             & 12.5 & 25.6 & 33.3 & - & - & - & - & - & -  \\
     & Text2Loc++ (Ours)                   & \overlaybold{\textbf{14.2}} & \overlaybold{\textbf{28.2}} & \overlaybold{\textbf{36.0}} & \overlaybold{\textbf{12.8}} & \overlaybold{\textbf{25.5}} & \overlaybold{\textbf{33.6}} & \overlaybold{\textbf{9.8}} & \overlaybold{\textbf{21.2}} & \overlaybold{\textbf{27.5}} \\
     \hdashline
     & Text2Pos~\cite{kolmet2022text2pos}  & 6.0 & 15.1 & 22.2 & 3.8 & 9.4 & 13.2 & 1.8 & 5.7 & 8.8 \\
Refinement     
     & Text2Loc~\cite{xia2024text2loc}     & 23.9 & 44.9 & 55.6 & 12.9 & 29.1 & 37.8 & 11.5 & 24.5 & 33.1 \\
     & Text2Loc++ (Ours)                   & \overlaybold{\textbf{28.4}} & \overlaybold{\textbf{51.6}} & \overlaybold{\textbf{62.5}} & \overlaybold{\textbf{27.2}} & \overlaybold{\textbf{49.9}} & \overlaybold{\textbf{60.3}} & \overlaybold{\textbf{18.5}} & \overlaybold{\textbf{36.7}} & \overlaybold{\textbf{46.4}} \\
    \midrule
    \multicolumn{11}{c}{Toronto~\cite{tan2020toronto3d}}\\
    \midrule
     & Text2Pos~\cite{kolmet2022text2pos}  & 2.1 & 6.3 & 9.6 & 1.9 & 5.7 & 8.7 & 0.8 & 2.1 & 3.4 \\    
     & Text2Loc~\cite{xia2024text2loc}     & 6.2 & 15.5 & 21.1 & 3.7 & 8.3 & 12.6 & 2.3 & 6.1 & 9.2 \\
Baseline
     & MambaPlace~\cite{shang2025mambaplace}     & 10.7 & 22.0 & 30.0 & - & - & - & - & - & - \\
     & CMMLoc~\cite{xu2025cmm}             & 7.6 & 17.9 & 24.1 & - & - & - & - & - & -  \\
     & Text2Loc++ (Ours)                   & \overlaybold{\textbf{13.2}} & \overlaybold{\textbf{26.4}} & \overlaybold{\textbf{34.6}} & \overlaybold{\textbf{13.2}} & \overlaybold{\textbf{25.7}} & \overlaybold{\textbf{33.8}} & \overlaybold{\textbf{10.6}} & \overlaybold{\textbf{21.3}} & \overlaybold{\textbf{27.4}} \\
    \hdashline
     & Text2Pos~\cite{kolmet2022text2pos}  & 3.4 & 10.0 & 14.7 & 1.5 & 4.6 & 6.9 & 1.2 & 3.5 & 5.3 \\
Refinement
     & Text2Loc~\cite{xia2024text2loc}     & 15.6 & 31.5 & 39.2 & 7.8 & 18.0 & 25.7 & 2.5 & 16.6 & 22.7 \\
     & Text2Loc++ (Ours)                   & \overlaybold{\textbf{23.1}} & \overlaybold{\textbf{41.3}} & \overlaybold{\textbf{49.2}} & \overlaybold{\textbf{21.1}} & \overlaybold{\textbf{38.5}} & \overlaybold{\textbf{47.7}} & \overlaybold{\textbf{14.5}} & \overlaybold{\textbf{28.4}} & \overlaybold{\textbf{35.0}} \\
    \midrule
    \multicolumn{11}{c}{KITTI360Pose (no color)~\cite{kolmet2022text2pos}} \\
    \midrule
     & Text2Pos~\cite{kolmet2022text2pos}  & 2.2 & 5.6 & 8.5 & 1.9 & 4.4 & 6.6 & 0.8 & 2.4 & 3.8 \\
Baseline 
     & Text2Loc~\cite{xia2024text2loc}     & 7.7 & 16.4 & 22.3 & 3.0 & 7.2 & 10.4 & 2.4 & 5.7 & 8.6 \\
     & Text2Loc++ (Ours)                   & \overlaybold{\textbf{10.6}} & \overlaybold{\textbf{20.8}} & \overlaybold{\textbf{27.1}} & \overlaybold{\textbf{10.2}} & \overlaybold{\textbf{19.7}} & \overlaybold{\textbf{25.9}} & \overlaybold{\textbf{7.0}} & \overlaybold{\textbf{14.9}} & \overlaybold{\textbf{20.1}} \\
     \hdashline
     & Text2Pos~\cite{kolmet2022text2pos}  & 2.0 & 5.6 & 8.3 & 1.2 & 3.5 & 5.7 & 0.9 & 2.4 & 3.9 \\
Refinement     
     & Text2Loc~\cite{xia2024text2loc}     & 7.7 & 16.7 & 22.7 & 3.1 & 7.2 & 10.4 & 2.4 & 6.1 & 9.2 \\
     & Text2Loc++ (Ours)                   & \overlaybold{\textbf{10.6}} & \overlaybold{\textbf{21.5}} & \overlaybold{\textbf{27.2}} & \overlaybold{\textbf{9.7}} & \overlaybold{\textbf{20.1}} & \overlaybold{\textbf{25.8}} & \overlaybold{\textbf{7.3}} & \overlaybold{\textbf{15.1}} & \overlaybold{\textbf{20.1}} \\
    \midrule
    \multicolumn{11}{c}{TUM~\cite{zhu2020tum}} \\
    \midrule
     & Text2Pos~\cite{kolmet2022text2pos}  & 5.4 & 14.3 & 21.2 & 7.0 & 15.7 & 20.8 & 2.1 & 9.4 & 13.9 \\
Baseline 
     & Text2Loc~\cite{xia2024text2loc}     & 15.5 & 29.8 & 39.5 & 5.5 & 14.6 & 22.2 & 7.7 & 16.4 & 24.3 \\
     & Text2Loc++ (Ours)                   & \overlaybold{\textbf{17.0}} & \overlaybold{\textbf{31.1}} & \overlaybold{\textbf{43.0}} & \overlaybold{\textbf{13.9}} & \overlaybold{\textbf{31.4}} & \overlaybold{\textbf{41.0}} & \overlaybold{\textbf{10.0}} & \overlaybold{\textbf{24.4}} & \overlaybold{\textbf{33.7}} \\
     \hdashline
     & Text2Pos~\cite{kolmet2022text2pos}  & 10.5 & 24.1 & 38.2 & 5.0 & 16.3 & 24.8 & 4.6 & 12.6 & 19.2 \\
Refinement     
     & Text2Loc~\cite{xia2024text2loc}     & 23.5 & 46.6 & 57.5 & 15.4 & 31.6 & 41.3 & 11.8 & 25.2 & 35.4 \\
     & Text2Loc++ (Ours)                   & \overlaybold{\textbf{27.2}} & \overlaybold{\textbf{50.8}} & \overlaybold{\textbf{62.5}} & \overlaybold{\textbf{26.2}} & \overlaybold{\textbf{48.4}} & \overlaybold{\textbf{60.0}} & \overlaybold{\textbf{23.8}} & \overlaybold{\textbf{41.9}} & \overlaybold{\textbf{53.3}} \\
    \midrule
    \multicolumn{11}{c}{Paris\_Lille~\cite{roynard2017parislille3d}}\\
    \midrule
     & Text2Pos~\cite{kolmet2022text2pos}  & 4.0 & 9.3 & 14.2 & 3.7 & 9.7 & 13.3 & 1.3 & 4.3 & 7.0 \\
Baseline     
     & Text2Loc~\cite{xia2024text2loc}     & 5.9 & \overlaybold{\textbf{14.8}} & 21.2 & 4.4 & 10.7 & 15.5 & 3.1 & 8.2 & 10.8 \\
     & Text2Loc++ (Ours)                   & \overlaybold{\textbf{6.9}} & 13.3 & \overlaybold{\textbf{21.5}} & \overlaybold{\textbf{6.2}} & \overlaybold{\textbf{15.9}} & \overlaybold{\textbf{22.0}} & \overlaybold{\textbf{6.5}} & \overlaybold{\textbf{14.1}} & \overlaybold{\textbf{19.2}} \\
    \hdashline
     & Text2Pos~\cite{kolmet2022text2pos}  & 4.7 & 12.8 & 19.8 & 4.4 & 12.2 & 17.3 & 1.9 & 6.9 & 11.2 \\
Refinement
     & Text2Loc~\cite{xia2024text2loc}     & 9.9 & 25.4& \overlaybold{\textbf{36.8}} & 7.6 & 18.5 & 25.9 & 6.3 & 16.7 & 25.4 \\
     & Text2Loc++ (Ours)                   & \overlaybold{\textbf{13.1}} & \overlaybold{\textbf{26.3}} & 35.6 & \overlaybold{\textbf{14.1}} & \overlaybold{\textbf{26.3}} & \overlaybold{\textbf{38.7}} & \overlaybold{\textbf{9.5}} & \overlaybold{\textbf{23.6}} & \overlaybold{\textbf{33.4}} \\
    \bottomrule
    \end{tabular}
   \caption{The evaluation results of the global place recognition. The "setting" column (baseline or refinement) indicates which datasets are used to train the network, as detailed in Table.~\ref{tab:dataset_split}. The terms simple, moderate, and complex refer to the linguistic complexity of the textual descriptions used during both training and testing. 
   }
    \label{tab:coarse_results}
    \vspace{-3mm}
\end{table*}

\begin{table*}[htp]
    \centering
    \scriptsize
    \fboxsep0.75pt
    \setlength{\extrarowheight}{0.2pt}
    \vspace{-1mm}
    \begin{tabular}{p{1.1cm} p{2.1cm} >{\fontsize{6.5pt}{6.5pt}\selectfont\centering\arraybackslash}m{1.1cm} 
                                      >{\fontsize{6.5pt}{6.5pt}\selectfont\centering\arraybackslash}m{1.1cm}
                                      >{\fontsize{6.5pt}{6.5pt}\selectfont\centering\arraybackslash}m{1.1cm} 
                                      >{\fontsize{6.5pt}{6.5pt}\selectfont\centering\arraybackslash}m{1.1cm} 
                                      >{\fontsize{6.5pt}{6.5pt}\selectfont\centering\arraybackslash}m{1.1cm} 
                                      >{\fontsize{6.5pt}{6.5pt}\selectfont\centering\arraybackslash}m{1.1cm} 
                                      >{\fontsize{6.5pt}{6.5pt}\selectfont\centering\arraybackslash}m{1.1cm} 
                                      >{\fontsize{6.5pt}{6.5pt}\selectfont\centering\arraybackslash}m{1.1cm} 
                                      >{\fontsize{6.5pt}{6.5pt}\selectfont\centering\arraybackslash}m{1.1cm}}
    \toprule
     &  & \multicolumn{9}{c}{Localization Recall (\%) - $\epsilon < 5/10/15m $  $\uparrow$} \\
    \cmidrule(lr){3-11}
     Setting & Method & \multicolumn{3}{c}{\textit{Simple}} & \multicolumn{3}{c}{\textit{Moderate}} & \multicolumn{3}{c}{\textit{Complex}} \\
     \cmidrule(lr){3-11}
     &  & k=1 & k=5 & k=10  & k=1 & k=5 & k=10 & k=1 & k=5 & k=10 \\
    \midrule
\multicolumn{11}{c}{KITTI360Pose~\cite{kolmet2022text2pos}} \\
    \midrule
     & Text2Pos~\cite{kolmet2022text2pos}  & 12/21/24 & 32/47/51 & 42/60/64 
                                           & 7/16/20 & 20/38/44 & 28/50/57
                                           & 3/6/8   & 9/19/24  & 14/29/36 \\

     & Text2Loc~\cite{xia2024text2loc}     & 34/47/50 & 61/74/76 & 71/83/85 
                                           & 28/39/43 & 54/67/70 & 65/77/80
                                           & 14/29/34 & 33/56/61 & 42/67/72 \\
Baseline    
     & MambaPlace~\cite{shang2025mambaplace} & 38/52/55 & 66/79/81 & 76/87/89 
                                           & - & - & -
                                           & - & - & - \\
     & CMMLoc~\cite{xu2025cmm}             & 39/53/56 & 67/80/82 & 77/87/89 
                                           & - & - & -
                                           & - & - & - \\
     & Text2Loc++ (Ours)                   & \overlaybold{\textbf{44}}/\overlaybold{\textbf{58}}/\overlaybold{\textbf{61}} & \overlaybold{\textbf{72}}/\overlaybold{\textbf{84}}/\overlaybold{\textbf{85}} & \overlaybold{\textbf{80}}/\overlaybold{\textbf{90}}/\overlaybold{\textbf{91}}
                                           & \overlaybold{\textbf{40}}/\overlaybold{\textbf{56}}/\overlaybold{\textbf{60}} & \overlaybold{\textbf{68}}/\overlaybold{\textbf{82}}/\overlaybold{\textbf{84}} & \overlaybold{\textbf{77}}/\overlaybold{\textbf{89}}/\overlaybold{\textbf{90}}
                                           & \overlaybold{\textbf{30}}/\overlaybold{\textbf{47}}/\overlaybold{\textbf{51}} & \overlaybold{\textbf{58}}/\overlaybold{\textbf{74}}/\overlaybold{\textbf{77}} & \overlaybold{\textbf{68}}/\overlaybold{\textbf{83}}/\overlaybold{\textbf{86}} \\
     \hdashline
     & Text2Pos~\cite{kolmet2022text2pos}  & 12/19/23 & 30/45/50 & 41/58/63 
                                           & 7/14/18  & 20/37/43 & 28/50/56
                                           & 3/6/9   & 9/20/25 & 15/30/37 \\
Refinement     
     & Text2Loc~\cite{xia2024text2loc}     & 35/48/51 & 64/76/78 & 73/84/85 
                                           & 28/40/49 & 56/68/71 & 67/78/80
                                           & 15/31/37 & 35/59/64 & 44/70/75 \\
     & Text2Loc++ (Ours)                   & \overlaybold{\textbf{40}}/\overlaybold{\textbf{56}}/\overlaybold{\textbf{60}} & \overlaybold{\textbf{67}}/\overlaybold{\textbf{82}}/\overlaybold{\textbf{84}} & \overlaybold{\textbf{76}}/\overlaybold{\textbf{89}}/\overlaybold{\textbf{90}} 
                                           & \overlaybold{\textbf{38}}/\overlaybold{\textbf{55}}/\overlaybold{\textbf{59}} & \overlaybold{\textbf{66}}/\overlaybold{\textbf{82}}/\overlaybold{\textbf{84}} & \overlaybold{\textbf{75}}/\overlaybold{\textbf{89}}/\overlaybold{\textbf{90}} 
                                           & \overlaybold{\textbf{27}}/\overlaybold{\textbf{45}}/\overlaybold{\textbf{50}} & \overlaybold{\textbf{54}}/\overlaybold{\textbf{72}}/\overlaybold{\textbf{76}} & \overlaybold{\textbf{64}}/\overlaybold{\textbf{82}}/\overlaybold{\textbf{84}} \\
    \midrule
\multicolumn{11}{c}{Paris\_CARLA~\cite{deschaud2021pariscarla3d}} \\
    \midrule
     & Text2Pos~\cite{kolmet2022text2pos}  & 3/7/11  & 11/23/31  & 18/35/44 
                                           & 1/3/5   & 4/11/17  & 8/18/28
                                           & 1/2/4   & 3/7/12   & 5/12/19 \\
                                           
     & Text2Loc~\cite{xia2024text2loc}     & 12/20/26 & 24/44/52 & 41/59/64 
                                           & 8/16/20  & 23/36/42 & 32/48/54
                                           & 2/5/7   & 7/16/23 & 12/25/33 \\
Baseline    
     & MambaPlace~\cite{shang2025mambaplace} & 11/22/27 & 31/50/56 & 39/61/68 
                                           & - & - & -
                                           & - & - & - \\
     & CMMLoc~\cite{xu2025cmm}             & 14/25/30 & 34/51/57 & 44/63/\overlaybold{\textbf{69}}
                                           & - & - & -
                                           & - & - & - \\
     & Text2Loc++ (Ours)                   & \overlaybold{\textbf{17}}/\overlaybold{\textbf{30}}/\overlaybold{\textbf{34}} & \overlaybold{\textbf{36}}/\overlaybold{\textbf{53}}/\overlaybold{\textbf{58}} & \overlaybold{\textbf{47}}/\overlaybold{\textbf{65}}/\overlaybold{\textbf{69}} 
                                           & \overlaybold{\textbf{14}}/\overlaybold{\textbf{26}}/\overlaybold{\textbf{30}}  & \overlaybold{\textbf{33}}/\overlaybold{\textbf{50}}/\overlaybold{\textbf{56}} & \overlaybold{\textbf{43}}/\overlaybold{\textbf{62}}/\overlaybold{\textbf{67}}
                                           & \overlaybold{\textbf{9}}/\overlaybold{\textbf{20}}/\overlaybold{\textbf{25}}  & \overlaybold{\textbf{24}}/\overlaybold{\textbf{43}}/\overlaybold{\textbf{49}}  & \overlaybold{\textbf{32}}/\overlaybold{\textbf{53}}/\overlaybold{\textbf{60}} \\
     \hdashline
     & Text2Pos~\cite{kolmet2022text2pos}  & 6/14/20  & 22/40/49 & 33/54/63 
                                           & 3/8/13   & 11/24/33 & 17/36/47
                                           & 1/5/8   & 7/18/27  & 13/29/41 \\
Refinement     
     & Text2Loc~\cite{xia2024text2loc}     & 25/43/49 & 52/74/79 & 64/85/88 
                                           & 16/30/36 & 39/60/66 & 52/73/78
                                           & 9/23/30  & 26/51/59 & 37/64/72 \\
     & Text2Loc++ (Ours)                   & \overlaybold{\textbf{33}}/\overlaybold{\textbf{51}}/\overlaybold{\textbf{55}} 
                                           & \overlaybold{\textbf{63}}/\overlaybold{\textbf{81}}/\overlaybold{\textbf{84}} 
                                           & \overlaybold{\textbf{75}}/\overlaybold{\textbf{90}}/\overlaybold{\textbf{92}}   
                                           & \overlaybold{\textbf{31}}/\overlaybold{\textbf{47}}/\overlaybold{\textbf{52}} 
                                           & \overlaybold{\textbf{60}}/\overlaybold{\textbf{78}}/\overlaybold{\textbf{82}} 
                                           & \overlaybold{\textbf{70}}/\overlaybold{\textbf{88}}/\overlaybold{\textbf{91}}    
                                           & \overlaybold{\textbf{18}}/\overlaybold{\textbf{34}}/\overlaybold{\textbf{41}} 
                                           & \overlaybold{\textbf{43}}/\overlaybold{\textbf{65}}/\overlaybold{\textbf{79}} 
                                           & \overlaybold{\textbf{55}}/\overlaybold{\textbf{76}}/\overlaybold{\textbf{82}} \\
    \midrule
\multicolumn{11}{c}{Toronto~\cite{tan2020toronto3d}}\\
    \midrule
     & Text2Pos~\cite{kolmet2022text2pos}  & 2/6/8   & 9/19/24  & 15/30/37 
                                           & 2/6/8   & 7/17/24  & 11/27/37
                                           & 1/2/4   & 3/7/12   & 5/13/21 \\
    
     & Text2Loc~\cite{xia2024text2loc}     & 10/16/19  & 24/39/44 & 34/54/59 
                                           & 5/11/14   & 17/27/33 & 25/39/47
                                           & 2/6/9    & 7/19/27  & 13/30/41 \\
Baseline    
     & MambaPlace~\cite{shang2025mambaplace} & 12/23/27 & 31/50/56 & 42/63/68 
                                           & - & - & -
                                           & - & - & - \\
     & CMMLoc~\cite{xu2025cmm}             & 9/17/22 & 28/43/50 & 41/59/65 
                                           & - & - & -
                                           & - & - & - \\
     & Text2Loc++ (Ours)                   & \overlaybold{\textbf{16}}/\overlaybold{\textbf{26}}/\overlaybold{\textbf{29}} & 
                                             \overlaybold{\textbf{36}}/\overlaybold{\textbf{54}}/\overlaybold{\textbf{59}} & \overlaybold{\textbf{47}}/\overlaybold{\textbf{68}}/\overlaybold{\textbf{73}}
                                           & \overlaybold{\textbf{14}}/\overlaybold{\textbf{25}}/\overlaybold{\textbf{28}} & \overlaybold{\textbf{32}}/\overlaybold{\textbf{49}}/\overlaybold{\textbf{54}} & \overlaybold{\textbf{42}}/\overlaybold{\textbf{62}}/\overlaybold{\textbf{67}}
                                           & \overlaybold{\textbf{9}}/\overlaybold{\textbf{21}}/\overlaybold{\textbf{25}} & \overlaybold{\textbf{24}}/\overlaybold{\textbf{43}}/\overlaybold{\textbf{49}} & \overlaybold{\textbf{34}}/\overlaybold{\textbf{56}}/\overlaybold{\textbf{62}} \\
    \hdashline
     & Text2Pos~\cite{kolmet2022text2pos}  & 4/9/13   & 15/28/35  & 23/41/50 
                                           & 1/4/7   & 6/15/22  & 11/25/33
                                           & 1/4/6   & 4/12/17  & 8/19/27 \\
Refinement
     & Text2Loc~\cite{xia2024text2loc}     & 18/30/35  & 39/59/64 & 52/72/76 
                                           & 10/19/23  & 28/45/50 & 39/59/66
                                           & 6/15/21   & 17/39/48 & 26/52/62 \\
     & Text2Loc++ (Ours)                   & \overlaybold{\textbf{30}}/\overlaybold{\textbf{43}}/\overlaybold{\textbf{46}} 
                                           & \overlaybold{\textbf{57}}/\overlaybold{\textbf{69}}/\overlaybold{\textbf{72}} 
                                           & \overlaybold{\textbf{68}}/\overlaybold{\textbf{81}}/\overlaybold{\textbf{84}} 
                                           & \overlaybold{\textbf{26}}/\overlaybold{\textbf{39}}/\overlaybold{\textbf{42}} 
                                           & \overlaybold{\textbf{51}}/\overlaybold{\textbf{67}}/\overlaybold{\textbf{71}} 
                                           & \overlaybold{\textbf{63}}/\overlaybold{\textbf{78}}/\overlaybold{\textbf{82}}
                                           & \overlaybold{\textbf{15}}/\overlaybold{\textbf{30}}/\overlaybold{\textbf{35}}  
                                           & \overlaybold{\textbf{36}}/\overlaybold{\textbf{57}}/\overlaybold{\textbf{62}}  
                                           & \overlaybold{\textbf{47}}/\overlaybold{\textbf{69}}/\overlaybold{\textbf{74}} \\
    \midrule
\multicolumn{11}{c}{KITTI360Pose (no color)~\cite{kolmet2022text2pos}} \\
    \midrule
     & Text2Pos~\cite{kolmet2022text2pos}  & 2/4/6   & 8/15/19  & 14/24/29 
                                           & 2/4/5   & 6/12/16  & 10/20/25
                                           & 1/2/3   & 3/8/11   & 6/13/18 \\
Baseline 
     & Text2Loc~\cite{xia2024text2loc}     & 8/12/13  & 23/32/35 & 33/45/48 
                                           & 3/5/7   & 11/17/21 & 17/26/30 
                                           & 2/4/5   & 7/14/17  & 11/21/26 \\
     & Text2Loc++ (Ours)                   & \overlaybold{\textbf{11}}/\overlaybold{\textbf{16}}/\overlaybold{\textbf{17}} & 
                                            \overlaybold{\textbf{26}}/\overlaybold{\textbf{38}}/\overlaybold{\textbf{41}} & \overlaybold{\textbf{36}}/\overlaybold{\textbf{51}}/\overlaybold{\textbf{54}} 
                                           & \overlaybold{\textbf{10}}/\overlaybold{\textbf{15}}/\overlaybold{\textbf{16}}  & \overlaybold{\textbf{25}}/\overlaybold{\textbf{36}}/\overlaybold{\textbf{39}} & \overlaybold{\textbf{34}}/\overlaybold{\textbf{49}}/\overlaybold{\textbf{52}}
                                           & \overlaybold{\textbf{6}}/\overlaybold{\textbf{11}}/\overlaybold{\textbf{12}}  & \overlaybold{\textbf{17}}/\overlaybold{\textbf{29}}/\overlaybold{\textbf{32}} & \overlaybold{\textbf{24}}/\overlaybold{\textbf{40}}/\overlaybold{\textbf{44}} \\
     \hdashline
     & Text2Pos~\cite{kolmet2022text2pos}  & 2/4/6   & 8/16/19  & 14/25/30
                                           & 1/2/4   & 5/10/14  & 8/16/22
                                           & 1/2/3   & 3/7/10   & 6/13/17 \\
Refinement     
     & Text2Loc~\cite{xia2024text2loc}     & 8/12/13  & 23/33/36 & 33/46/50 
                                           & 4/6/7   & 12/18/21 & 18/28/32 
                                           & 2/4/6   & 8/15/19  & 12/23/28 \\
     & Text2Loc++ (Ours)                   & \overlaybold{\textbf{10}}/\overlaybold{\textbf{16}}/\overlaybold{\textbf{17}} & \overlaybold{\textbf{26}}/\overlaybold{\textbf{38}}/\overlaybold{\textbf{41}} & \overlaybold{\textbf{36}}/\overlaybold{\textbf{51}}/\overlaybold{\textbf{54}} 
                                           & \overlaybold{\textbf{10}}/\overlaybold{\textbf{15}}/\overlaybold{\textbf{16}} & \overlaybold{\textbf{24}}/\overlaybold{\textbf{36}}/\overlaybold{\textbf{39}} & \overlaybold{\textbf{33}}/\overlaybold{\textbf{49}}/\overlaybold{\textbf{52}}
                                           & \overlaybold{\textbf{6}}/\overlaybold{\textbf{11}}/\overlaybold{\textbf{12}}  & \overlaybold{\textbf{17}}/\overlaybold{\textbf{29}}/\overlaybold{\textbf{32}} & \overlaybold{\textbf{24}}/\overlaybold{\textbf{40}}/\overlaybold{\textbf{44}} \\
    \midrule
\multicolumn{11}{c}{TUM~\cite{zhu2020tum}} \\
    \midrule
     & Text2Pos~\cite{kolmet2022text2pos}  & 5/14/20 & 19/41/55 & 29/57/73 
                                           & 6/13/20 & 16/34/47 & 25/50/61
                                           & 1/6/13  & 10/24/41 & 17/37/56 \\
Baseline 
     & Text2Loc~\cite{xia2024text2loc}     & 11/29/40 & 29/58/68 & 42/74/83 
                                           & 6/14/23  & 16/37/49 & 25/49/64 
                                           & 5/13/23  & 18/40/55 & 29/58/72 \\
     & Text2Loc++ (Ours)                   & \overlaybold{\textbf{12}}/\overlaybold{\textbf{30}}/\overlaybold{\textbf{39}} & 
                                             \overlaybold{\textbf{31}}/\overlaybold{\textbf{59}}/\overlaybold{\textbf{68}} & \overlaybold{\textbf{47}}/\overlaybold{\textbf{76}}/\overlaybold{\textbf{83}}
                                           & \overlaybold{\textbf{9}}/\overlaybold{\textbf{28}}/\overlaybold{\textbf{35}} & \overlaybold{\textbf{26}}/\overlaybold{\textbf{59}}/\overlaybold{\textbf{65}} & \overlaybold{\textbf{39}}/\overlaybold{\textbf{74}}/\overlaybold{\textbf{82}}
                                           & \overlaybold{\textbf{8}}/\overlaybold{\textbf{19}}/\overlaybold{\textbf{26}}  & \overlaybold{\textbf{26}}/\overlaybold{\textbf{51}}/\overlaybold{\textbf{59}} & \overlaybold{\textbf{40}}/\overlaybold{\textbf{68}}/\overlaybold{\textbf{76}} \\
     \hdashline
     & Text2Pos~\cite{kolmet2022text2pos}  & 10/21/29 & 28/55/67 & 39/70/81
                                           & 4/11/20  & 19/42/57 & 29/58/72
                                           & 4/11/19  & 15/35/49 & 22/49/62 \\
Refinement     
     & Text2Loc~\cite{xia2024text2loc}     & 25/44/52 & 54/79/83 & 68/89/91
                                           & 15/32/41 & 37/66/74 & 52/80/86
                                           & 9/21/32  & 30/54/64 & 41/70/80 \\
     & Text2Loc++ (Ours)                   & \overlaybold{\textbf{29}}/\overlaybold{\textbf{47}}/\overlaybold{\textbf{54}} & 
                                             \overlaybold{\textbf{57}}/\overlaybold{\textbf{79}}/\overlaybold{\textbf{82}}& \overlaybold{\textbf{61}}/\overlaybold{\textbf{88}}/\overlaybold{\textbf{90}}
                                           & \overlaybold{\textbf{21}}/\overlaybold{\textbf{47}}/\overlaybold{\textbf{53}} & \overlaybold{\textbf{49}}/\overlaybold{\textbf{77}}/\overlaybold{\textbf{81}} & \overlaybold{\textbf{63}}/\overlaybold{\textbf{87}}/\overlaybold{\textbf{90}}
                                           & \overlaybold{\textbf{20}}/\overlaybold{\textbf{38}}/\overlaybold{\textbf{46}} & \overlaybold{\textbf{42}}/\overlaybold{\textbf{69}}/\overlaybold{\textbf{76}} & \overlaybold{\textbf{54}}/\overlaybold{\textbf{82}}/\overlaybold{\textbf{87}} \\
    \midrule
\multicolumn{11}{c}{Paris\_Lille~\cite{roynard2017parislille3d}}\\
    \midrule
     & Text2Pos~\cite{kolmet2022text2pos}  & 3/9/14   & 12/30/41  & 21/47/61 
                                           & 3/10/15  & 10/28/40  & 18/43/55
                                           & 1/4/8    & 5/16/25   & 12/29/41 \\
Baseline     
     & Text2Loc~\cite{xia2024text2loc}     & \overlaybold{\textbf{5}}/12/\overlaybold{\textbf{18}} & 16/\overlaybold{\textbf{36}}/\overlaybold{\textbf{46}} & 28/\overlaybold{\textbf{55}}/\overlaybold{\textbf{65}}
                                           & \overlaybold{\textbf{4}}/\overlaybold{\textbf{11}}/\overlaybold{\textbf{17}} & 12/28/39  & 20/44/59 
                                           & 2/5/9    & 9/21/31   & 14/35/48 \\
     & Text2Loc++ (Ours)                   & \overlaybold{\textbf{5}}/\overlaybold{\textbf{13}}/\overlaybold{\textbf{18}}  & \overlaybold{\textbf{18}}/34/\overlaybold{\textbf{46}} & \overlaybold{\textbf{30}}/53/\overlaybold{\textbf{65}}
                                           & \overlaybold{\textbf{4}}/\overlaybold{\textbf{11}}/\overlaybold{\textbf{17}} & \overlaybold{\textbf{16}}/\overlaybold{\textbf{36}}/\overlaybold{\textbf{48}} & \overlaybold{\textbf{26}}/\overlaybold{\textbf{55}}/\overlaybold{\textbf{68}}
                                           & \overlaybold{\textbf{5}}/\overlaybold{\textbf{10}}/\overlaybold{\textbf{15}} & \overlaybold{\textbf{15}}/\overlaybold{\textbf{33}}/\overlaybold{\textbf{43}} & \overlaybold{\textbf{25}}/\overlaybold{\textbf{50}}/\overlaybold{\textbf{64}} \\
    \hdashline
     & Text2Pos~\cite{kolmet2022text2pos}  & 5/11/17  & 16/54/45  & 28/53/65
                                           & 4/10/15  & 12/32/43  & 21/50/63
                                           & 1/7/12   & 8/26/36   & 17/42/53 \\
Refinement
     & Text2Loc~\cite{xia2024text2loc}     & 10/21/\overlaybold{\textbf{27}} & 33/50/\overlaybold{\textbf{57}}   & 47/67/74
                                           & 8/14/21  & 24/40/51  & 39/61/71 
                                           & 5/13/20  & 19/38/50  & 29/56/68 \\
     & Text2Loc++ (Ours)                   & \overlaybold{\textbf{12}}/\overlaybold{\textbf{22}}/26 & \overlaybold{\textbf{35}}/\overlaybold{\textbf{51}}/\overlaybold{\textbf{57}} & \overlaybold{\textbf{51}}/\overlaybold{\textbf{70}}/\overlaybold{\textbf{77}} 
                                           & \overlaybold{\textbf{11}}/\overlaybold{\textbf{22}}/\overlaybold{\textbf{25}} & \overlaybold{\textbf{34}}/\overlaybold{\textbf{53}}/\overlaybold{\textbf{59}} & \overlaybold{\textbf{48}}/\overlaybold{\textbf{71}}/\overlaybold{\textbf{77}} 
                                           & \overlaybold{\textbf{8}}/\overlaybold{\textbf{16}}/\overlaybold{\textbf{20}} & \overlaybold{\textbf{27}}/\overlaybold{\textbf{48}}/\overlaybold{\textbf{58}} & \overlaybold{\textbf{38}}/\overlaybold{\textbf{69}}/\overlaybold{\textbf{77}} \\
    \bottomrule
    \end{tabular}
    \caption{The evaluation results of the fine localization. The "setting" column (baseline or refinement) indicates which datasets are used to train the network, as detailed in Table.~\ref{tab:dataset_split}. The terms simple, moderate, and complex refer to the linguistic complexity of the textual descriptions used during both training and testing. 
    }
    \label{tab:fine_results}
    \vspace{-4mm}
\end{table*}

\subsection{Evaluation Criteria} 
Following~\cite{kolmet2022text2pos}, we use Retrieve Recall at top-$k$ ($k \in \{1, 3, 5\}$) to evaluate text-submap global place recognition. The ground-truth submap is defined as the submap whose centroid coordinates are closest to the ground-truth location among all candidates. For a given text description, if the ground-truth submap appears within the top-$k$ predicted submaps, the case is considered positive; otherwise, it is negative.
For assessing localization performance, we evaluate with respect to the top $k$ retrieved candidates ($k \in \{1, 5, 10\}$) and report localization recall. Localization recall measures the proportion of successfully localized queries if their error falls below specific error thresholds, specifically $\epsilon < 5/10/15m$ by default.
Table~\ref{tab:coarse_results} and ~\ref{tab:fine_results} shows the accuracy results for global place recognition and fine localization respectively.

\subsection{Results}
We compare our proposed Text2Loc++ with several state-of-the-art methods: Text2Pos~\cite{kolmet2022text2pos}, RET~\cite{wang2023text}, MambaPlace~\cite{shang2025mambaplace}, and CMMLoc~\cite{xu2025cmm}. For completeness, we also include our previous version, Text2Loc~\cite{xia2024text2loc}, to highlight improvements.
For a fair comparison, all methods are evaluated on six benchmark datasets for both global place recognition and fine-grained localization.  
We evaluate under two training settings: baseline and refinement, as detailed in Section~\ref{dataset:data_splitting} and Table~\ref{tab:dataset_split}. In addition, we analyze the model’s performance under different levels of text complexity to assess robustness to linguistic variation.

\subsubsection{Global Place Recognition}
\noindent\textbf{Baseline Networks.} 
Text2Loc++ achieves the best performance on the KITTI360Pose test set, reaching a recall of 35.3\%  at top-1. Notably, this outperforms the recall achieved by the current state-of-the-art method CMMLoc by a wide margin of \textbf{9.3\%}. Furthermore, Text2Loc++ achieves recall rates of 57.6\% and 66.9\% at top-3 and top-5, respectively, representing substantial improvements of 7.8\% and 6.0\% relative to the performance of CMMLoc. 
These improvements demonstrate the efficiency of our Text2Loc++ to capture local information from the cross-model and generate more discriminative global descriptors. More qualitative results are given in Section~\ref{sec: vis}. 
A horizontal comparison between ``color'' and ``no-color'' input for the KITTI360Pose dataset shows that all models experience a significant performance drop, highlighting the importance of color information in the 3D localization task.

\noindent\textbf{3D Generalization.} To evaluate the generalization ability of our network, 
we train solely on the KITTI360Pose dataset and evaluate on other datasets, including Paris\_CARLA~\cite{deschaud2021pariscarla3d}, Toronto~\cite{tan2020toronto3d}, TUM~\cite{zhu2020tum} and Paris~\cite{roynard2017parislille3d} datasets.
Excitingly, our network performs this task remarkably well. Across different datasets, we achieve strong results in top-1/3/5 recall, consistently outperforming all other models. Moreover, we observe that with color input, the model becomes more sensitive to variations in point cloud configurations, leading to larger performance differences across datasets.
\\ \noindent\textbf{Refinement Network.} 
To further enhance generalization, we extend training beyond KITTI360Pose by incorporating additional datasets—Paris\_CARLA (with color), and Lille and TUM (without color). As shown in Table~\ref{tab:coarse_results}, this refinement strategy significantly improves performance on previously unseen datasets, particularly Toronto (color) and Lille (no\_color), demonstrating the effectiveness of our approach.
Despite using only a small amount of additional data, Text2Loc++ achieves strong generalization across diverse domains in LiDAR-based place recognition. Notably, the inclusion of extra training data does not noticeably degrade performance on the original KITTI360Pose test set. These results highlight the robustness of our model and suggest that, with the availability of more diverse LiDAR datasets, further performance gains can be achieved—ultimately contributing to more consistent localization across varied environments.
\\ \noindent\textbf{Text Generalization.} 
To assess the model’s generalizability from a language perspective, we evaluate its performance across three levels of text complexity: simple, moderate, and complex. On the KITTI360Pose test set, the performance with moderate text input remains comparable to that with simple descriptions, suggesting that the model can effectively handle moderately complex language. However, performance degrades with complex inputs, indicating that challenges remain in filtering redundant information, extracting relevant content, and processing longer descriptions.
On other datasets, the performance drop is more pronounced as text complexity increases. This suggests that the difficulty of cross-modal alignment in unfamiliar domains further limits the model’s ability to interpret and leverage complex language.

\begin{table}[t]
    \centering
    \renewcommand{\arraystretch}{1.2}
    \setlength{\tabcolsep}{8pt}

    \begin{tabular}{lccc}
        \toprule
        \multirow{2}{*}{\centering \textbf{Setting} \rule{0pt}{13pt}} &  
        \multicolumn{3}{c}{\scriptsize{\textbf{Submap Retrieval Recall (\%) $\uparrow$} }} \\
        \cmidrule(lr){2-4}
         & $k=1$ & $k=3$ & $k=5$ \\ 
        \midrule
        w/o masked instance & 32.1 & 53.5 & 62.6 \\ 
        matching  & 10.1 & 21.3 & 28.5 \\
       \midrule
       Text2Loc++ & \overlaybold{\textbf{35.3}} & \overlaybold{\textbf{57.6}} & \overlaybold{\textbf{66.9}} \\
 \bottomrule
    \end{tabular}
    \caption{Performance analysis of the masked instance training. "matching" refers to retaining only the instances in the submaps that are explicitly described in the text instead of masked instance training.}
    \label{tab:masked_analysis}
\end{table}

\subsubsection{Fine Localization} 
To enhance localization accuracy, prior works~\cite{kolmet2022text2pos, wang2023text} introduce a fine localization stage. For a fair comparison, we adopt the same training configuration as in~\cite{kolmet2022text2pos, wang2023text} for our fine localization network. As shown in Table~\ref{tab:fine_results}, we report the top-$k$ ($k = 1, 5, 10$) recall rates under different error thresholds $\epsilon < 5/10/15,\text{m}$.
On the KITTI360Pose (color) test set with simple text descriptions, Text2Loc++ achieves a top-1 recall rate of 44\% under an error bound of $\epsilon < 5,\text{m}$, outperforming the previous state-of-the-art CMMLoc by \textbf{15\%}. It is observed that the application of MCL and MIT during training leads to a substantial performance gain, with TextLoc++ achieving an approximate 30\% increase compared to Text2Loc. Besides, we evaluate the performance gain from global place recognition to fine localization ($\epsilon < 5m$). While Text2Loc shows a modest improvement from 29\% to 34\% (5\% increase), Text2Loc++ achieves a more substantial gain of 10\%. This performance margin serves as a direct indicator of the model's fine-grained localization capability. Our results demonstrate that the integration of MIT in fine localization effectively enhances multi-modal information fusion, leading to superior precision during the fine localization stage. Moreover, Text2Loc++ consistently maintains superior performance when the localization error tolerance is relaxed or when $k$ increases. These results demonstrate that Text2Loc++ can more accurately interpret textual descriptions and achieve stronger semantic understanding of point clouds compared to prior methods. 
Text2Loc++  also achieves the best performance in the other datasets and refinement training. 
For different levels of text difficulty, moderate text descriptions show a slight performance drop compared to simple descriptions. Since fine localization requires more precise extraction of textual information than global place recognition, it is more sensitive to the complexity of the input text. With complex text input, the model struggles to accurately match the textual descriptions with the corresponding information in the submaps, leading to a noticeable decline in overall performance.

\section{Ablation Study}
\label{sec: discussions}

\subsection{Ablation Study of Masked Instance Training}

In Table~\ref {tab:masked_analysis}, the removal of MIT strategy results in an 8\% decrease in model performance. Additionally, the complete exclusion of instances that not aligned with the textual descriptions substantially impairs the model's generalization capability. These observations underscore the efficacy of masked instance training in mitigating the challenges associated with many-to-many cross-modal alignment during the training process. MIT effectively strikes a balance between accuracy and generalization performance in such scenario.

\subsection{Ablation Study of Modality-aware Contrastive Learning}
\begin{table}[tb]
    \centering
    \renewcommand{\arraystretch}{1.2}
    \setlength{\tabcolsep}{8pt}

    \begin{tabular}{lccc}
        \toprule
        \multirow{2}{*}{\centering \textbf{Setting} \rule{0pt}{13pt}} &  
        \multicolumn{3}{c}{\scriptsize{\textbf{Submap Retrieval Recall (\%) $\uparrow$} }} \\
        \cmidrule(lr){2-4}
         & $k=1$ & $k=3$ & $k=5$ \\ 
        \midrule
        w/o instance loss & 33.2 & 54.9 & 64.0 \\ 
        w/o text loss & 33.4 & 56.2 & 65.3 \\ 
        w/o submap loss & 32.5 & 54.1 & 63.6 \\ 
       \midrule
       Text2Loc++ & \overlaybold{\textbf{35.3}} & \overlaybold{\textbf{57.6}} & \overlaybold{\textbf{66.9}} \\
 \bottomrule
    \end{tabular}
    \caption{Performance analysis of the each component of the proposed modality-aware contrastive learning.}
    \label{tab:aux_loss_analysis}
\end{table}

Table.~\ref{tab:aux_loss_analysis} presents an ablation study evaluating the impact of different losses for the modality-aware contrastive learning. We could conclude that removing any loss component leads to a drop in performance across all recall levels from 2\% to 3\% in top-1 Recall, highlighting the importance of each component to the model’s overall effectiveness. Among the proposed components, the submap loss exerts the most significant influence on the overall performance. This underscores the critical importance of learning discriminative representations within the embedding space; specifically, ensuring that distinct submaps are mapped to their appropriate, well-separated embeddings is the most significant for robust cross-modal alignment.

The accuracy gains from MIT and MCL show that these components effectively suppress the false pull in Fig.~\ref{fig:coarse2} caused by shared text–submap pairs in training and mitigate the resulting ambiguity.

\subsection{Ablation Study of Text Distillation}
\label{sec:multiple_analysis}


In Table~\ref {tab:coarse_group_alignment}, we report Text2Loc++ performance regarding text distillation. In both moderate and complex texts, we observe that disabling text distillation results in a decrease of 11\% and 7\% in top-5 performance, respectively. This highlights the critical role of text group alignment. Additionally, we find that modifying other modules cannot compensate for this loss. In the moderate setting, we attempted to increase the complexity of text during training by using linguistically diverse expressions conveying the same meaning. However, this approach did not lead to any performance improvement. This indicates that in the case of moderate text, the increased complexity prevents us from directly constructing a multimodal latent space. Therefore, the task must be divided: first, by constructing the latent space based on simple text, and then by using text distillation to map the complex text into the pre-established latent space.

\begin{table}[t]
  \centering
  \small
  \begin{tabular}{p{1.0cm} >{\centering\arraybackslash}m{0.25cm} >
  {\centering\arraybackslash}m{1.9cm} >
  {\centering\arraybackslash}m{0.9cm} >
  {\centering\arraybackslash}m{0.9cm} >
  {\centering\arraybackslash}m{0.9cm}}
    \toprule
     \multirow{2}{*}{\centering \textbf{Level} \rule{0pt}{13pt}} & 
     \multirow{2}{*}{\centering \textbf{TD} \rule{0pt}{13pt}} & 
     \multirow{2}{*}{\centering \textbf{Extra} \rule{0pt}{13pt}} &
        \multicolumn{3}{c}{\footnotesize{\textbf{Submap Retrieval Recall (\%) $\uparrow$} }} \\
     \cmidrule(lr){4-6}
      & & & $k=1$ & $k=3$ & $k=5$ \\
    \midrule
    \multirow{3}{*}{\footnotesize{Moderate}} 
     & \xmark & - & 22.7 & 41.8 & 50.8 \\
     & \xmark & \footnotesize{Shuffle} & 23.1 & 41.7 & 50.5 \\
     & \cmark & - & \overlaybold{\textbf{34.0}} & \overlaybold{\textbf{56.8}} & \overlaybold{\textbf{65.4}} \\
    \midrule
    \multirow{5}{*}{\footnotesize{Complex}}  
     & \xmark & - & 19.0 & 35.5 & 44.1 \\
     & \xmark & \footnotesize{Translate} & 16.7 & 31.8 & 39.8 \\
     & \xmark & \scriptsize{Trans.-LoRA*} & \overlaybold{\textbf{29.9}} & \overlaybold{\textbf{50.5}} & \overlaybold{\textbf{59.5}} \\
     & \cmark & \footnotesize{w/o LoRA} & 24.8 & 44.6 & 53.8\\
     & \cmark & - & \underline{27.9} & \underline{48.1} & \underline{57.5} \\
    \bottomrule
  \end{tabular}
  \caption{Comparison of whether using text distillation for moderate and complex text on the KITTI360Pose benchmark. 
  \textbf{TD} means text distillation. Several additional settings are applied during both training and testing phases. "Shuffle" involves randomizing the format of moderate descriptions. For a specific position, we utilize various formats of descriptions during training. "Translate" denotes the use of ChatGPT-4o~\cite{openai2024gpt4ocard} to simplify complex text descriptions before feeding them into the language encoder for both training and testing.}
  \label{tab:coarse_group_alignment}
\end{table}

As shown in Table~\ref{tab:coarse_results}, even with the incorporation of MIT, MCL, and TD, complex text input consistently yields approximately 10\% lower performance than the simple input. As linguistic structures become more entangled and blended—with non-parallel syntax and the inclusion of irrelevant or extraneous content—the model exhibits difficulty in capturing all useful information. Although we initially consider applying text simplification via ChatGPT-4o~\cite{openai2024gpt4ocard} to mitigate this issue, empirical results show no performance gain. A likely explanation is that the simplification process introduces semantic loss, which adversely affects model predictions. As discussed in Section~\ref{sec: robust}, the model demonstrates high sensitivity to sentence-level content, where even minor omissions or distortions result in notable performance drops. Conversely, finetuning the T5 module with the LoRA method significantly enhances accuracy in the first stage. In the case of complex text descriptions, the best results are achieved using the translate-LoRA approach, where the finetuned T5 model first translates complex text into simplified text, which is then passed into our model. However, this method is approximately 10 times slower ($\sim$ 300 ms) than directly using the T5 encoder with LoRA ($\sim$ 30 ms). Given the minimal performance difference and the substantial efficiency gain, we adopt the direct T5 encoder approach in our final model.

\begin{table}[t]
    \centering
    \renewcommand{\arraystretch}{1.2}
    \setlength{\tabcolsep}{8pt}

    \begin{tabular}{llccc}
        \toprule
        \multirow{2}{*}{\centering \textbf{Parameter} \rule{0pt}{13pt}} & \multirow{2}{*}{\centering \textbf{Value} \rule{0pt}{13pt}} & 
        \multicolumn{3}{c}{\scriptsize{\textbf{Submap Retrieval Recall (\%) $\uparrow$} }} \\
        \cmidrule(lr){3-5}
         & & $k=1$ & $k=3$ & $k=5$ \\ 
        \midrule
        Loss & pairwise & 31.5 & 53.3 & 62.5 \\
        \midrule
        \multirow{3}{*}{Latent Space}  & dim = 64 & 28.3 & 48.5 & 57.4 \\
                                       & dim = 128 & 32.9 & 54.8 & 64.1 \\
                                       & dim = 512 & \overlaybold{\textbf{35.6}} & \overlaybold{\textbf{58.3}} & \overlaybold{\textbf{67.9}} \\
        \midrule
        \multirow{4}{*}{HTM Layers} &  no HTM & 31.0 & 51.0 & 59.8 \\
                                          & 1 \; + \; 2 & 34.7 & 56.7 & 65.9 \\
                                          & 2 \; + \; 1 & 34.9 & 57.6 & \underline{66.9}\\
                                          & 2 \; + \; 2 & 33.8 & 56.1 & 65.5 \\      
        \midrule
        \multirow{2}{*}{Pooling} & Intra Mean & 35.0 & 57.5 & 66.4 \\
                                 & Inter Mean & 32.0 & 53.5 & 62.9 \\
        \midrule
        \multirow{5}{*}{\shortstack{Backbones \\ (Text)}} & T5\_base & 34.1 & 56.8 & 65.5 \\
                                   & T5\_large & 34.0 & 56.7 & 65.8 \\
                                   & Llama3.2-1B & 33.2 & 54.6 & 63.7 \\
                                   & Llama3.2-3B & 33.0 & 55.5 & 64.7 \\
                                   & CLIP-B/16 & 33.2 & 54.6 & 63.7 \\ 
        \midrule
        \multirow{3}{*}{Input Branches} & no number & 33.6 & 54.6 & 63.7 \\
                                      & no color & 26.0 & 45.6 & 54.8 \\
                                      & no position & 21.0 & 40.1 & 49.8 \\
       \midrule
       Default Config & - & \underline{35.3} & \underline{57.6} & \underline{66.9} \\
 \bottomrule
    \end{tabular}
    \caption{Performance analysis of global place recognition in the different configurations in the simple text mode. The default config utilize InfoNCE loss\cite{oord2019}, 256 latent space, HTM (intra = 1 $+$ inter = 1), max pooling in intra and inter encoder, T5\_large \cite{2020t5} backbones in the text module, and number, color, position branches.}
    \label{tab:coarse_ablation_study}
\end{table}

\subsection{Analysis of Network Settings}
\label{sec: ablation}
The following ablation studies evaluate the effectiveness of different
 components of Text2Loc, including both the text-submap global place recognition and fine localization. 

{\bf Global place recognition.} In Table.~\ref{tab:coarse_ablation_study}, we assess the relative contribution of modules and settings. 

We replace the InfoNCE loss \cite{oord2019} by pairwise ranking loss \cite{kiros2014} that causes a slight decrease of the perforamce.
    
By systematically varying the latent space dimension from 64 to 512, we find that when it is below 128, changes in dimensionality substantially influence the results. However, once the dimension exceeds 128, further increases have only a minor effect on performance. To ensure a fair comparison with other models, we set the dimension to 256 as our default setting.
    
We also explore the effect of hierarchical transformer modules (HTM). Removing the HTM layers altogether results in lower performance, whereas stacking layers (e.g., 1+2 or 2+1) introduces a minor alteration on the recall at all values.

We further study different pooling mechanisms for aggregating features, comparing intra- and inter-mean with the max pooling.  The results suggest that intra-mean pooling performs on par with the default setting, while inter-mean pooling falls short. 

Then, we investigate how different backbones affect the final results. Overall, we found that changing the backbone does not significantly alter submap retrieval recall. We tested an encoder-only T5 (since the T5 decoder component was not used in our model) and a decoder-only Llama. Even though Llama’s architecture is more complex, it failed to improve performance on our task. The underlying reason is that our task requires precise extraction of linguistic information rather than broad understanding of multiple languages, and thus different LLM architectures performed similarly in this regard. Furthermore, we used the language encoder from CLIP \cite{radford2021learning} and found that the multimodal text encoder does not outperform a pure language model in our task.

 Finally, we analyze the role of different input branches. Removing each branch leads to a noticeable drop in overall performance, emphasizing the importance of preserving all input modalities.

      Our default configuration yields the highest recall, reflecting a balanced choice of loss function, latent dimension, HTM configuration, pooling strategy, backbone architecture, and input branches.

\begin{table}[t]
    \centering
    \footnotesize
    \resizebox{0.47\textwidth}{!}{%
    \begin{tblr}{
      colspec = {p{2.4cm} *{3}{>{\centering\arraybackslash}m{1cm}} p{0.02cm}},
      cells = {c},
      cell{1}{2} = {c=3}{}, 
      hline{1,9} = {-}{0.08em},
      hline{2} = {2-4}{0.03em},
      hline{3,8} = {-}{0.05em},
    }
                    & Localization Recall (\%) - $\epsilon < 5m$ $\uparrow$ &  &  &  \\
    Method          & $k=1$         & $k=5$         & $k=10$    &    \\
    Text2Pos~\cite{kolmet2022text2pos}       
                    & 12.2           & 31.9          & 43.0     &    \\                    
    Text2Pos*       & 36.6            & 64.7          & 72.8    &    \\
    Text2Loc++\_PMC    & 36.8           & 64.6          & 72.9    &     \\
    Text2Loc++\_CCAT   & 39.9           & 69.7          & 78.2    &     \\
    Text2Loc++\_P+C   & 41.4           & 70.6          & 79.1    &     \\    
    Text2Loc++ (Ours) & \overlaybold{\textbf{43.8}}   & \overlaybold{\textbf{72.1}} & \overlaybold{\textbf{80.0}}   & \\
    \end{tblr}}
    \caption{Ablation study of the fine localization on the KITTI360Pose benchmark. * indicates the fine localization network from Text2Pose~\cite{kolmet2022text2pos}, and the submaps retrieved through our global place recognition. Text2Loc++\_CCAT indicates the removal of only the PMC while retaining the CCAT in our network. Conversely, Text2Loc++\_PMC keeps the PMC but replaces the CCAT with the text-instance matcher in Text2Pos.  Text2Loc++\_P+C includes both PMC and CCAT, but excludes the MIT during training.}
    \label{tab:fine_ablation_study}
\end{table}

\subsection{Ablation Study of Fine Localization}
To analyze the effectiveness of each proposed module in our matching-free fine-grained localization, we separately evaluate the Cascaded Cross-Attention Transformer (CCAT) and Prototype-based Map Cloning (PMC) module, denoted as Text2Loc++\_CCAT and Text2Loc++\_PMC. For a fair comparison, all methods utilize the same submaps retrieved from our global place recognition.  The results are shown in Table.~\ref{tab:fine_ablation_study}. Text2Pos* significantly outperforms the origin results of Text2Pos~\cite{kolmet2022text2pos}, indicating the superiority of our proposed global place recognition.  Notably, replacing the matcher in Text2Pos~\cite{kolmet2022text2pos} with our CCAT results in about 10\% improvements at top-1 on the test set. 
We also observe the inferior performance of Text2Loc++\_PMC to the proposed method when interpreting only the proposed PMC module into the Text2Pos~\cite{kolmet2022text2pos} fine localization network. The results are consistent with our expectations since PMC can lead to the loss of object instances in certain submaps. Combining both modules achieves better performance, improving the performance by 10\% at top-1 on the test set. This demonstrates adding more training submaps by PMC is beneficial for our matching-free strategy without any text-instance matches. The inclusion of Masked Instance Training (MIT) leads to a consistent performance gain across all metrics, with a notable improvement in the most stringent accuracy thresholds. This demonstrates that by forcing the model to focus on task-relevant spatial signals and ignore environmental noise, MIT effectively enhances the robustness and precision of the learned representations for fine-grained localization.


\begin{table}[t]
    \centering
    \resizebox{0.47\textwidth}{!}{
    \begin{tblr}{
      cells = {c},
      hline{1,4} = {-}{0.15em},
      hline{2} = {1-3}{0.15em},
      hline{2} = {4}{},
    }
    Method         & Parameters (M) & Runtime (ms)  & Localization Recall \\
    Text2Loc++\_Matcher     & 2.08           & 43.11         & 0.34                \\
    Text2Loc++ (Ours) & \textbf{1.06}  & \textbf{2.27} & \textbf{0.40}       
    \end{tblr}}
    \caption{Computational cost requirement analysis of our fine localization network on the KITTI360Pose test dataset. }
    \label{tab: efficiency analysis}
\end{table}

\section{Discussion}
\subsection{Computational Cost Analysis}
\label{sec: eff}
In this section, we analyze the computational requirements of our two stage networks in terms of parameter count and time efficiency.
For a fair comparison, all methods are evaluated on the KITTI360Pose test set using a single NVIDIA TITAN X (12 GB) GPU.
In global recognition, Text2Loc++ requires \SI{22.75}{ms} and \SI{12.37}{ms} to generate global descriptors for a textual query and a submap, respectively, while Text2Pos~\cite{kolmet2022text2pos} achieves this in \SI{2.31}{ms} and \SI{11.87}{ms}. The longer inference time for Text2Loc++ stems from the additional frozen T5 (\SI{30.21}{ms} with LoRA and \SI{21.18}{ms} without LoRA) and HTM modules (\SI{1.57}{ms}). Our text and 3D encoders contain \SI{13.65}{M} (excluding T5) and \SI{1.84}{M} parameters, respectively.
For fine localization, we substitute the proposed matching-free CCAT module with the text–instance matcher from \cite{kolmet2022text2pos, wang2023text}, denoted as Text2Loc++\_Matcher.
As shown in Table~\ref{tab: efficiency analysis}, Text2Loc++ is nearly twice as parameter-efficient as previous works~\cite{kolmet2022text2pos, wang2023text} and requires only 5\% of their inference time. This gain arises primarily because previous approaches employ SuperGlue~\cite{sarlin20superglue} as a matcher, which is computationally heavy and time-consuming.
Furthermore, our matching-free design eliminates the need for the Sinkhorn algorithm~\cite{NIPS2013_af21d0c9}, further improving efficiency without sacrificing performance.


\begin{figure*}[t]
    \centering
    \includegraphics[width=0.92\textwidth]{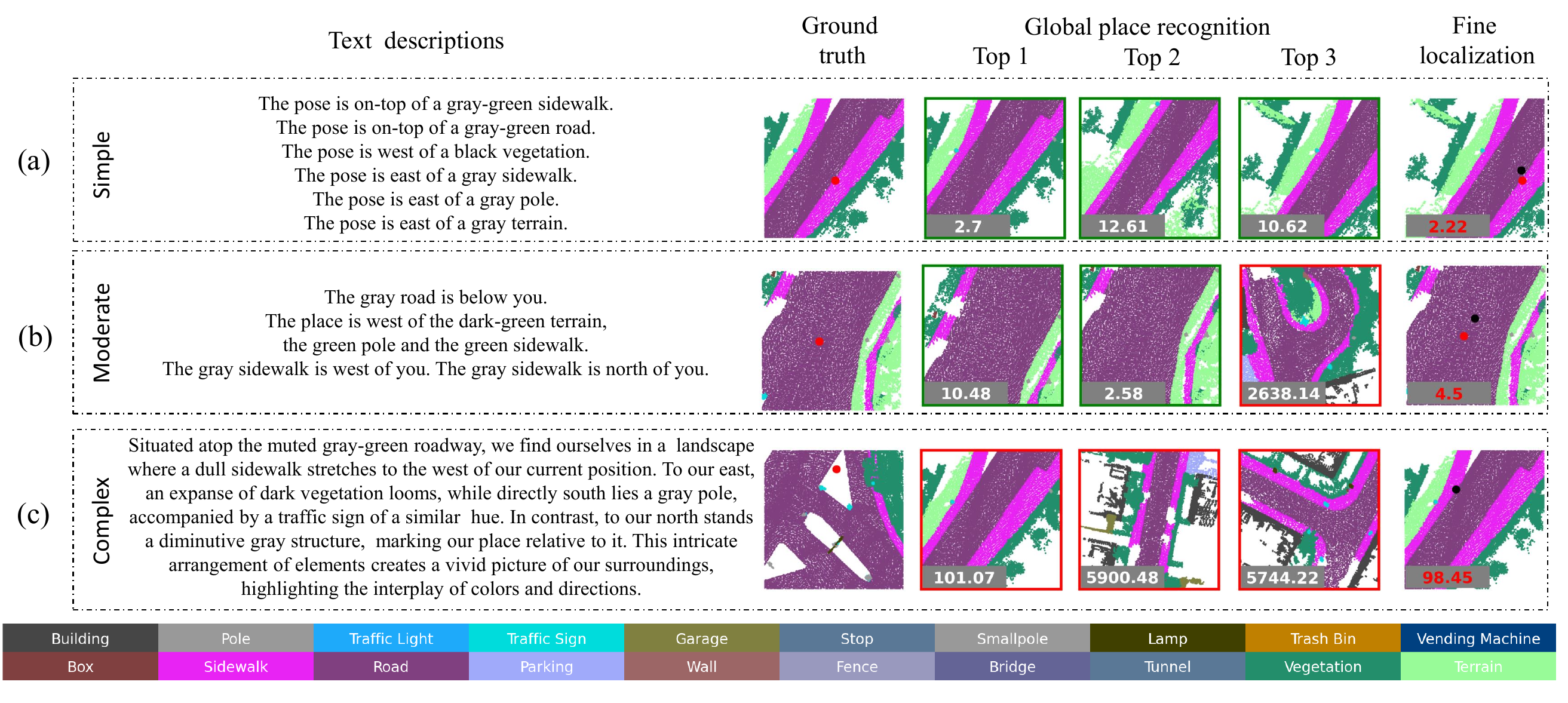}
    \caption{Qualitative localization results on the KITTI360Pose dataset: In global place recognition, the numbers in top3 retrieval submaps represent center distances between retrieved submaps and the ground truth. Green boxes indicate positive submaps containing the target location, while red boxes signify negative submaps. For fine localization, red and black dots represent the ground truth and predicted target locations, with the red number indicating the distance between them. (a), (b), and (c) uses simple, moderate, and complex text descriptions respectively.
    }
    \label{fig:qualit_results}
    \vspace{-0.5em}
\end{figure*}

\subsection{Qualitative Analysis}
\label{sec: vis}

To complement the quantitative analysis, Fig.~\ref{fig:qualit_results} presents qualitative examples, showing two correct localizations and one failure case from text descriptions.
Given a query text, we visualize the ground truth, the top-3 retrieved submaps, and the fine localization predictions. Within the text–submap global place recognition setting, a retrieved submap is defined as positive if it includes the target location. We provide text inputs with different levels of complexity across various scenes. Subfigures (a)–(c) correspond to simple, moderate, and complex text descriptions, respectively. The results allow us to intuitively observe how the model performs under different degrees of textual complexity.
In most cases, Text2Loc++ effectively retrieves the ground-truth submap or spatially adjacent ones. Nonetheless, as shown in (b), some negative submaps may still appear among the top-3 retrieved candidates.
By comparing (a) and (b), Text2Loc++ demonstrates its ability to predict more accurate locations by leveraging positively retrieved submaps during fine localization.
With increasing text complexity, more failure cases like (c) occur, where all retrieved submaps are negative. In these cases, fine localization has difficulty predicting accurate positions, indicating its reliance on coarse localization.

Through the qualitative results obtained from text queries of varying complexity, we identify two key challenges in outdoor multimodal text-to-point-cloud retrieval.
(1) Scene similarity. Negative submaps may contain instances visually similar to the ground truth due to the limited diversity of outdoor environments. This highlights the necessity of highly discriminative representations to effectively distinguish between submaps.
(2) Text complexity. Highly complex text descriptions, even when containing sufficient location cues, can degrade localization performance because of their intricate linguistic structures and the presence of irrelevant information.



\subsection{Robustness Analysis}
\label{sec: robust}

\begin{figure}[t]
    \centering
    \includegraphics[width=0.5\textwidth]{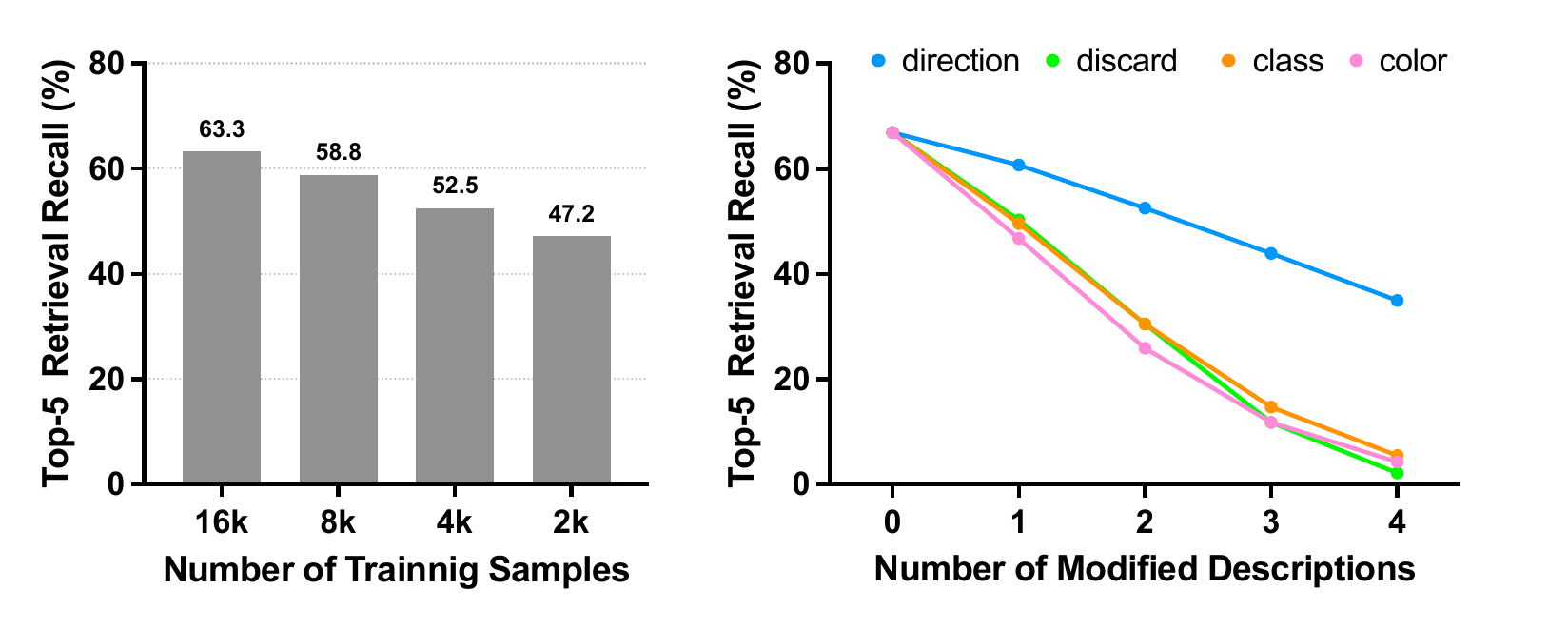}

    \caption{Robustness analysis comparing different metrics.}
    \label{fig:robust_analysis}
    \vspace{-1.5em}
\end{figure}

In this section, we investigate the impact of text modifications on localization accuracy. To provide a comprehensive analysis, we systematically alter specific components of the query text—namely color, direction, and semantic class—at varying levels of severity. Specifically, we incrementally modify between 1 and 5 sentences, ranging from minor to major changes. We also evaluate the network’s robustness to erroneous versus missing information by randomly omitting sentences. Details of the modification strategies are provided in Table~\ref{tab:robust_change_instruction}, and all evaluations are conducted on the KITTI360Pose test set. The top-5 results are illustrated in Fig.~\ref{fig:robust_analysis}.

\begin{table} [t]
    \centering
    \begin{tabular}{c c}
        \toprule
        \textbf{Change Type} & \textbf{Hint} \\
        \midrule
        \textbf{Original} &  The pose is east of a gray road. \\ 
        \midrule
        \textbf{Color} & The pose is east of a \underline{green} road. \\
    
        \textbf{Direction} & The pose is \underline{west} of a gray road. \\
    
        \textbf{Semantic class} &  The pose is east of a gray \underline{sidewalk}. \\
    
        \textbf{Discard} & \textit{(Sentence deleted)} \\
    
        \bottomrule
    \end{tabular}
    \caption{Change Instruction}
    \label{tab:robust_change_instruction}
    \vspace{-1.5em}
\end{table}

Our findings show that modifying or removing even a single sentence can cause a notable performance drop, especially for color and semantic-class cues. Altering one sentence in these categories leads to an approximately 30\% decrease in both top-1 and top-5 recall, indicating that the model is highly sensitive to contextual text cues. In contrast, direction-related modifications have the weakest impact.

The degradation patterns further differ across the types: directional changes lead to a near-linear decline as more sentences are modified, whereas color and class modifications show an exponential drop. Except for direction, erroneous information has an effect comparable to sentence omission, suggesting that incorrect context can be as harmful as missing information. These results highlight the need for robust context modeling in text-based localization.
We further evaluate robustness under reduced training data following the protocol in \cite{zha2023rank}, decreasing the training set from 16k to 2k samples. As shown in Fig.~\ref{fig:robust_analysis}, top-5 recall decreases smoothly from 63.3\% to 47.2\%, showing that the model retains strong retrieval performance under substantial data scarcity.

\subsection{Embedding Space Analysis}
\label{sec: embedding space}

\begin{figure}[t]
    \centering
    \includegraphics[width=0.9\columnwidth]{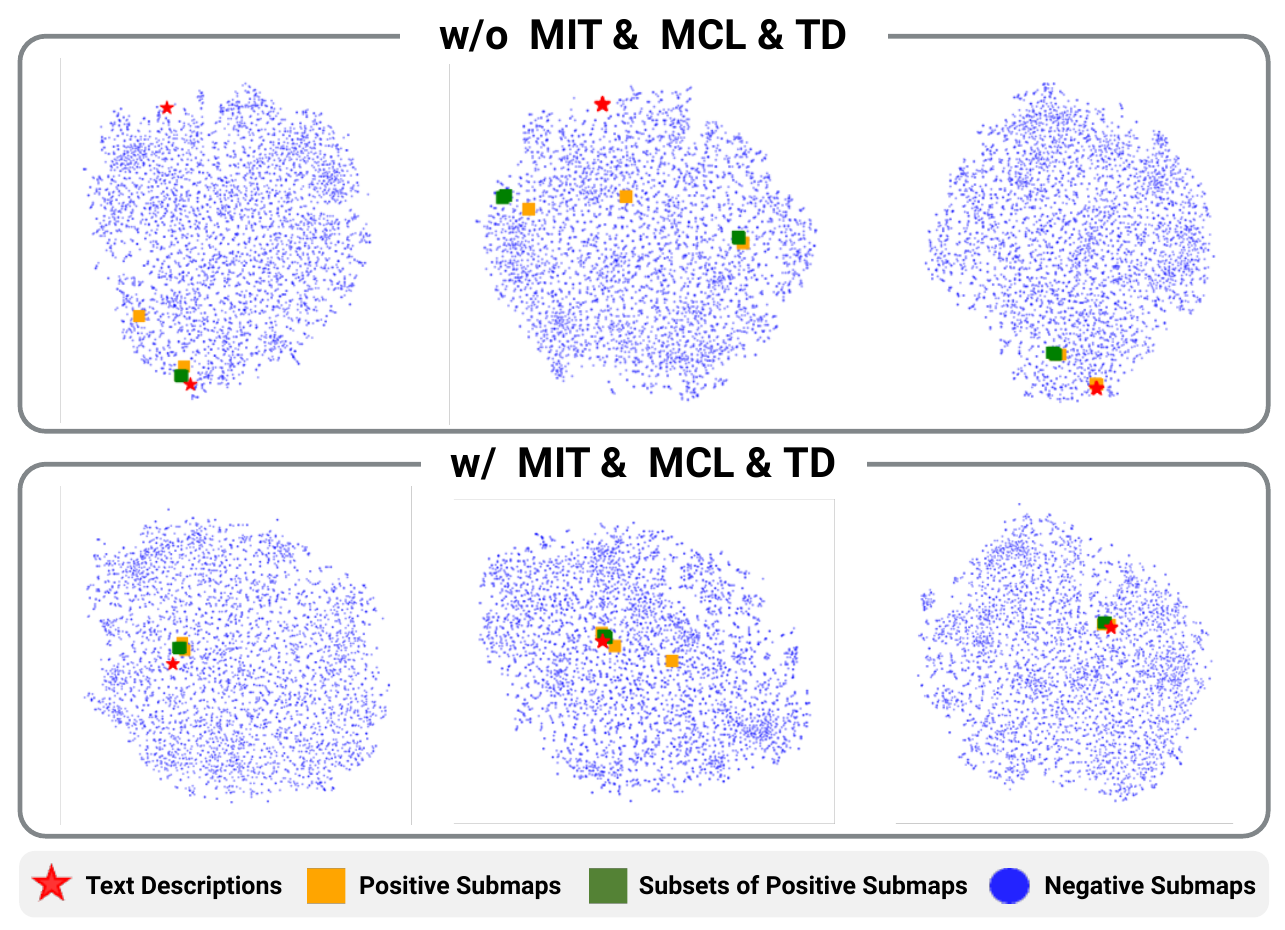}
    \caption{T-SNE illustration comparison between with and without masked instance training (MIT), modality-aware  contrastive learning (MCL), and text distillation (TD).}
    \label{fig:tsne}    
    \vspace{-1.5em}
\end{figure}

To illustrate the structure of the learned embedding space, we visualize it using T-SNE~\cite{van2008visualizing}, as presented in Fig.~\ref{fig:tsne}. The baseline method Text2Loc~\cite{xia2024text2loc}, not using masked instance training (MIT), modality-aware contrastive learning (MCL), and the text distillation (TD), results in a less discriminative embedding space, where positive submaps tend to lie far from their corresponding query text descriptions and are often dispersed across the space. At the same time, they also cause text embeddings with different expressions but the same meaning, as well as subsets of positive submaps (submaps containing only a subset of instances from the original submaps), to have a large distance from the original submaps. This negatively affects our ability to accurately perform text-based submap retrieval during testing. In contrast, the method with MIT, MCL, TD brings positive submaps, subset of positive submaps, and query text representations significantly closer together within the embedding distance. It shows that the proposed network indeed results in a more discriminative cross-model space for recognizing places.

\section{Conclusion}
In this work, we address the under-explored problem of localizing 3D point cloud submaps from natural language descriptions, a key step toward more intuitive and flexible human-robot spatial interaction.
We propose Text2Loc++, a novel coarse-to-fine framework that captures fine-grained semantic context across modalities. For global place recognition, we introduce an attention-based language encoder and a modality-aware contrastive learning strategy with masked instance training to better align text and submaps.
To handle complex language inputs, we incorporate text distillation and LoRA-based tuning. Furthermore, we are the first to propose a matching-free fine localization design, which is lighter, faster, and more accurate.
Extensive experiments demonstrate that Text2Loc++ outperforms prior methods and generalizes well across diverse linguistic and geometric inputs.
We hope this work inspires future research in language-grounded 3D localization and human-centric spatial understanding.
 
\vspace{0.2cm}
\noindent
{\bf Acknowledgments.} This work was supported by the Anhui Provincial Natural Science Foundation (2508085MF142), by the New Generation Artificial Intelligence-National Science and Technology Major Project (Grant No. 2025ZD0122603), by the ERC Advanced Grant SIMULACRON, and by the Royal Academy of Engineering (RF\textbackslash 201819\textbackslash 18\textbackslash 163). Some figures created by https://BioRender.com.




\ifCLASSOPTIONcaptionsoff
  \newpage
\fi



%


{
\small
\bibliographystyle{IEEEtran}
\bibliography{reference}
}

%
\newpage
\begin{IEEEbiography}[{\includegraphics[width=1in,height=1.20in,clip,keepaspectratio]{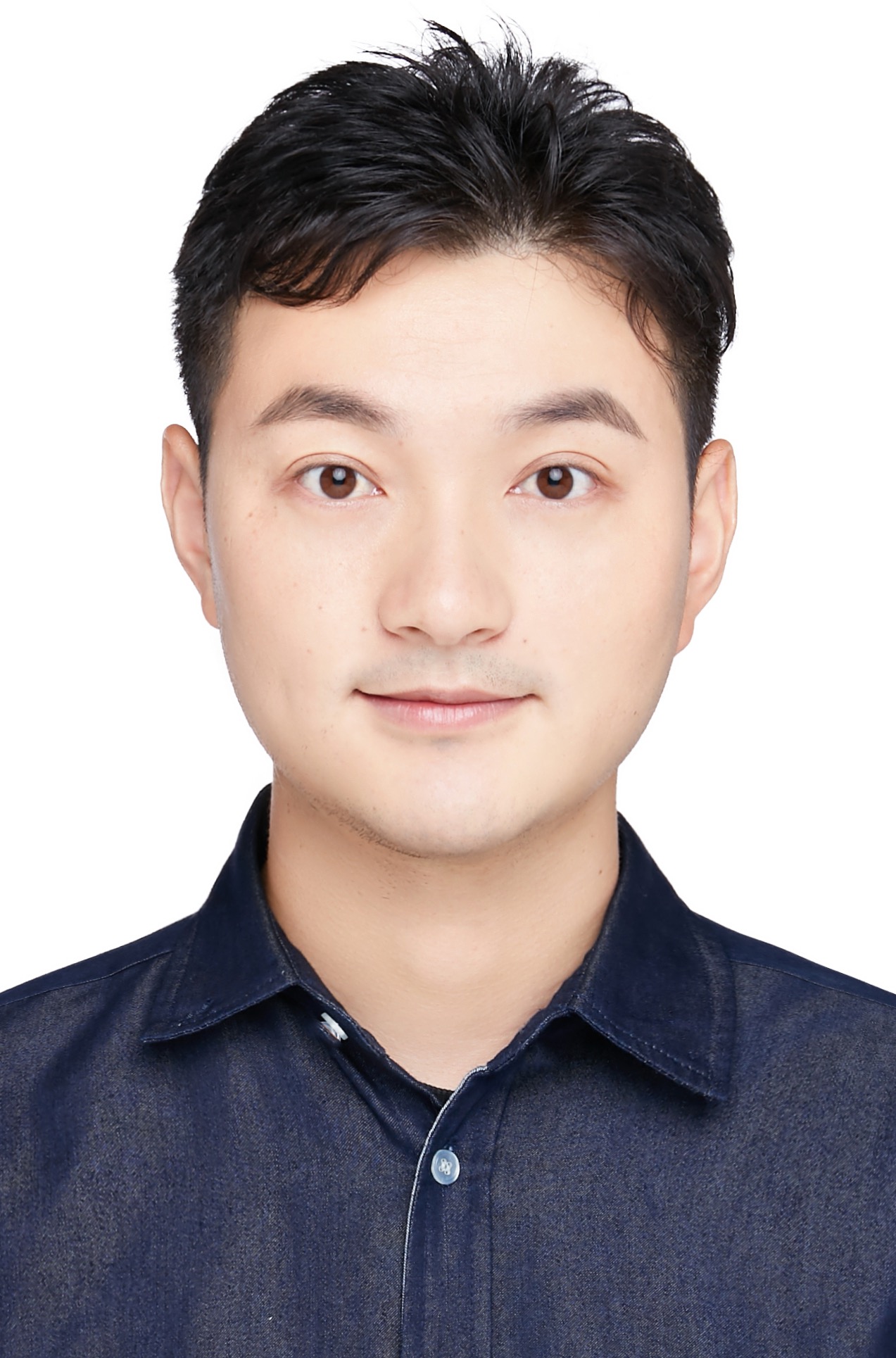}}]{Yan Xia} is now a Professor at University of Science and Technology of China (USTC). He was a senior researcher in the Chair of Computer Vision and Artificial Intelligence at Technical University of Munich (TUM) and a research scientist at Munich Center for Machine Learning (MCML), working with Prof. Daniel Cremers.  He obtained his PhD degree from TUM in 2023 and was a visiting scholar in Visual Geometry Group (VGG) at University of Oxford. His research interests include 3D vision and autonomous driving.
\end{IEEEbiography}

\begin{IEEEbiography}[{\includegraphics[width=1in,height=1.20in,clip,keepaspectratio]{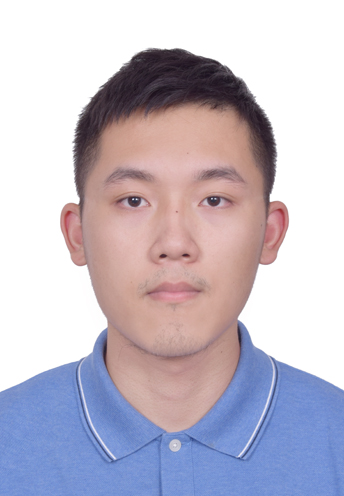}}]{Letian Shi} is currently a researcher in the Institute for Computer Science, Artificial Intelligence
and Technology (INSAIT). He obtained his M.Sc. degree in Data Engineering and Analytics from TUM in 2023. His research interests include 3D vision, multimodal model, and 3D generation.
\end{IEEEbiography}


\begin{IEEEbiography}[{\includegraphics[width=1in,height=1.20in,clip,keepaspectratio]{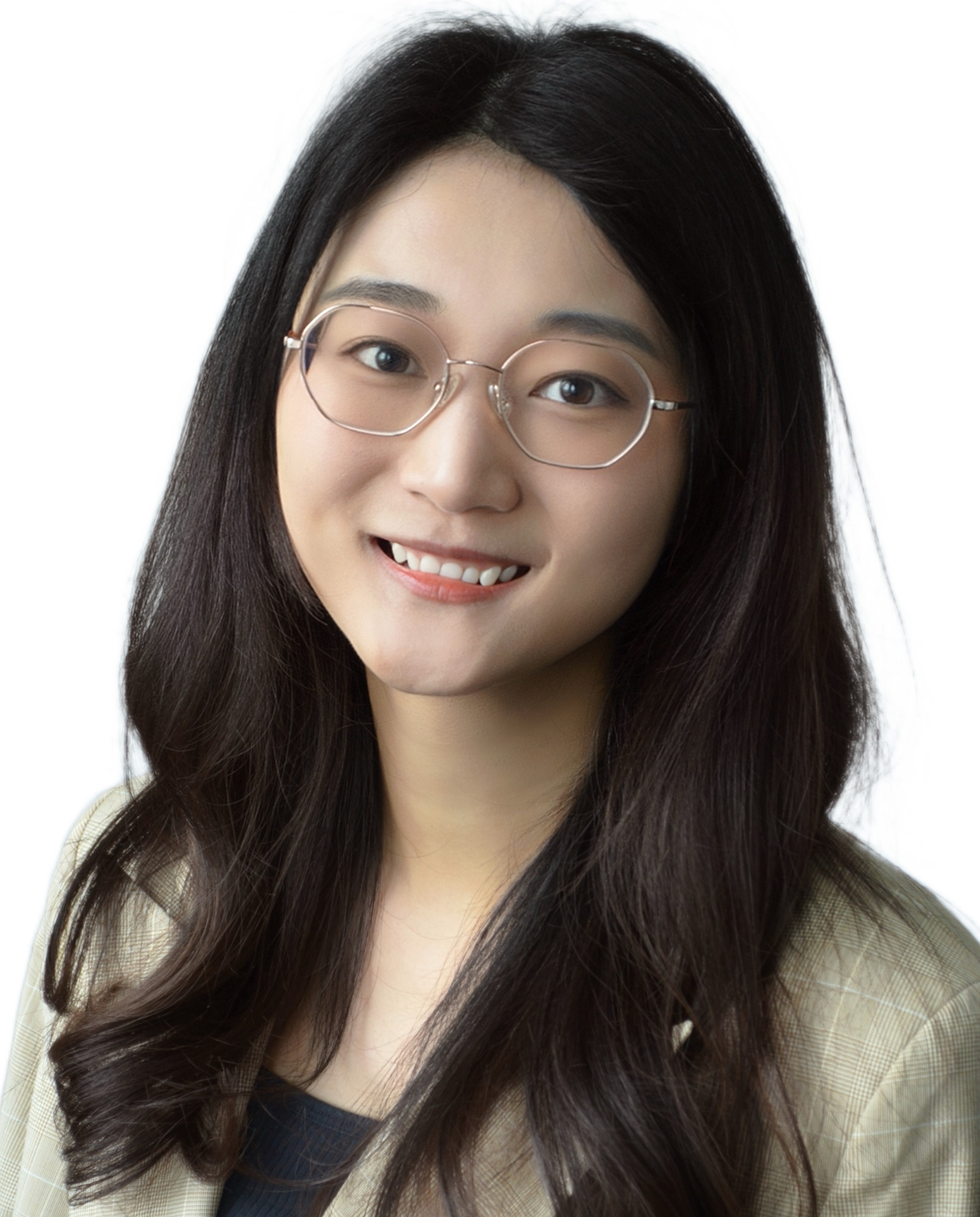}}]{Yilin Di} is currently a PhD researcher at the Technical University of Munich (TUM), working under the supervision of Prof. Henrik Semb. She previously worked as a research scientist at the Helmholtz Zentrum München (HMGU). She obtained her master’s degree from TUM in 2021. Her research interests include 3D cell models, organoid development, and live imaging.
\end{IEEEbiography}

\begin{IEEEbiography}
[{\includegraphics[width=1in,height=1.20in,clip,keepaspectratio]{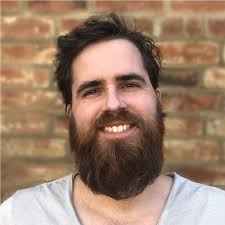}}]{João~F.~Henriques} is a Research Fellow of the Royal Academy of Engineering, working at the Visual Geometry Group (VGG) at the University of Oxford. His research focuses on computer vision and deep learning, with the goal of making machines more perceptive, intelligent and capable of helping people. He created the KCF and SiameseFC visual object trackers, which won the highly competitive VOT Challenge twice, and are widely deployed in consumer hardware, from Facebook apps to commercial drones. 
His research spans many topics: robot mapping and navigation, including reinforcement learning and 3D geometry; multi-agent cooperation and "friendly" AI; as well as various forms of learning, from self-supervised, causal, and meta-learning, including optimization theory. 
\end{IEEEbiography}

\begin{IEEEbiography}
[{\includegraphics[width=1in,height=1.20in,clip,keepaspectratio]{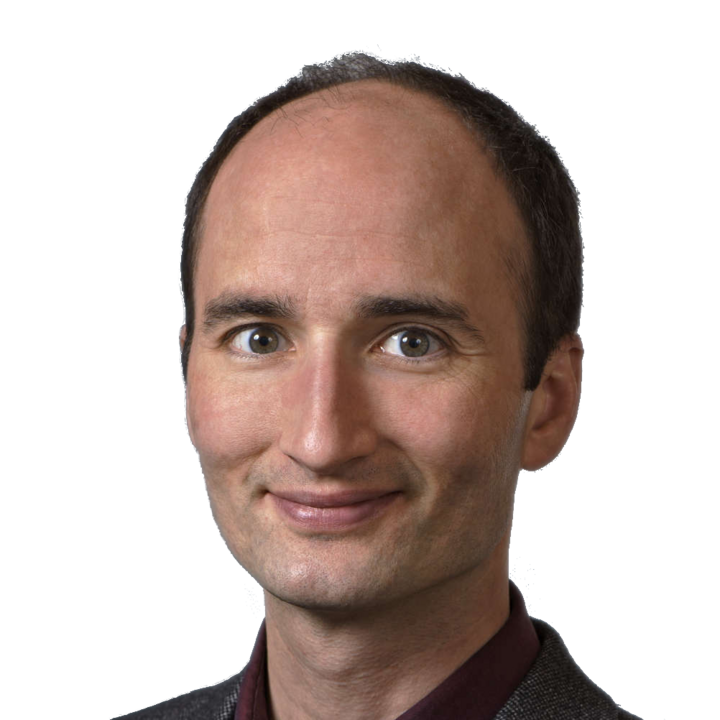}}]{Daniel~Cremers}
is a Professor at TUM, where he is holding the Chair of Computer Vision and Artificial Intelligence. He is also a co-founder of Artisense, a deep-tech startup developing computer vision and AI solutions for robotics and autonomous driving. Daniel has served as an area chair for ICCV, ECCV, CVPR, ACCV, IROS, etc., and as a program chair for ACCV 2014. In 2018, he was an organizer of ECCV in Munich. His publications received several awards and have been cited more than 70000 times. In 2016, Daniel received the Leibniz Award, the biggest award in German academia.
\end{IEEEbiography}
    



\end{document}